\documentclass[10pt,twocolumn,letterpaper]{article}

\usepackage{iccv}
\usepackage{times}
\usepackage{epsfig}
\usepackage{graphicx}
\usepackage{amsmath}
\usepackage{subfigure}
\usepackage{amssymb}

\usepackage{amsfonts}
\usepackage{amsmath,bm}
\usepackage{multirow}

\usepackage{rotating}

\usepackage[linesnumbered,ruled,vlined]{algorithm2e}
\usepackage{algpseudocode}  
 
\usepackage[pagebackref=true,breaklinks=true,letterpaper=true,colorlinks,bookmarks=false]{hyperref}

\iccvfinalcopy 

\def\iccvPaperID{8617} 

\ificcvfinal\fi

\begin{document}

\title{Top-Related Meta-Learning Method for Few-Shot Object Detection}

\author{First Author\\
Institution1\\
Institution1 address\\
{\tt\small firstauthor@i1.org}
\and
Second Author\\
Institution2\\
First line of institution2 address\\
{\tt\small secondauthor@i2.org}
}
\author{
  Qian Li \thanks{https://github.com/futureisatyourhand}\\
  State Key Laboratory of Computer Architecture, Institute of Computing Technology,\\
University of Chinese Academy of Sciences, Beijing, China\\
  \texttt{liqian18s@ict.ac.cn} \\
  Nan Guo\\
 State Key Laboratory of Computer Architecture, Institute of Computing Technology,Beijing, China\\
   \texttt{guonan@ict.ac.cn} \\
   Xiaochun Ye, Duo Wang, Dongrui Fan and Zhimin Tang\\
  State Key Laboratory of Computer Architecture, Institute of Computing Technology,\\
University of Chinese Academy of Sciences, Beijing, China \\
\texttt{\{yexiaochun,wangduo18z,fandr,tang\}@ict.ac.cn}
}

\maketitle
\ificcvfinal\thispagestyle{empty}\fi

\begin{abstract}
Many meta-learning methods are proposed for few-shot detection. However, previous most methods have two main problems, poor detection APs, and strong bias because of imbalance and insufficient datasets. Previous works mainly alleviate these issues by additional datasets, multi-relation attention mechanisms and sub-modules. However, they require more cost. In this work, for meta-learning, we find that the main challenges focus on related or irrelevant semantic features between categories. Therefore, based on semantic features, we propose a Top-C classification loss (i.e., TCL-C) for classification task and a category-based grouping mechanism for category-based meta-features obtained by the meta-model. The TCL-C exploits the true-label prediction and the most likely C-1 false classification predictions to improve detection performance on few-shot classes. According to similar appearance (i.e., visual appearance, shape, and limbs etc.) and environment in which objects often appear, the category-based grouping mechanism splits categories into disjoint groups to make similar semantic features more compact between categories within a group and obtain more significant difference between groups, alleviating the strong bias problem and further improving detection APs. The whole training consists of the base model and the fine-tuning phases. According to grouping mechanism, we group the meta-features vectors obtained by meta-model, so that the distribution difference between groups is obvious, and the one within each group is less. Extensive experiments on Pascal VOC dataset demonstrate that ours which combines the TCL-C with category-based grouping significantly outperforms previous state-of-the-art methods for few-shot detection. Compared with previous competitive baseline, ours improves detection APs by almost 4\% for few-shot detection.
\end{abstract}

\section{Introduction}
Recently, neural network has progressed quickly for computer vision. Various efficient methods \cite{Redmon2015You} \cite{Redmon2017YOLO9000} \cite{tian2019fcos} \cite{kong2019foveabox} depend on large labeled datasets. However, when datasets are insufficient, it may result in overfitting and hurting generalization performance. On the contrary, there is a quite difference between the human vision system and the computer vision system. For the unlabeled datasets, the human vision system can classify, locate and describe. Computer systems cannot do those. Despite most state-of-the-art methods can succeed, they require more expensive datasets that are the labeled with auxiliary descriptions, such as shape, scene or color etc.

The predecessors propose few-shot learning methods \cite{wertheimer2019few-shot} \cite{lifchitz2019dense} \cite{ShahTask}, solving the above issues, and few-shot learning includes classification, detection and segementation. Few-shot detection \cite{bansal2018zero-shot} \cite{PorikliZero} \cite{SaligramaZero} is one of the most challenging tasks. This paper finds two main challenges. First, due to just few samples, the features which are extracted from standard CNNs are not suitable for few-shot learning, directly. In previous most state-of-the-art few-shot learning methods, the classification is often regarded as the standard task. For each iteration of training, classification is a binary classification task for YOLOv2 \cite{Redmon2017YOLO9000}, resulting in bias problem and hurting performance on the other classes. Then, many methods \cite{PinheiroAdaptive} \cite{XieDual} \cite{HebertWatch} are proposed for auxiliary features related to description. However, it is difficult to ensure whether the external datasets are beneficial and tell which is noise. Therefore, many methods \cite{PinheiroAdaptive} \cite{ZemelIncremental} \cite{XieDual} learn auxiliary features by sub-modules to improve performance, requiring  more labeled datasets and more parameters.

\begin{figure*} 
 \scriptsize
\subfigure[The visualization of category grouping.]
{
  \begin{minipage}[c]{0.5\textwidth} 
  \includegraphics[width=1.0\textwidth]{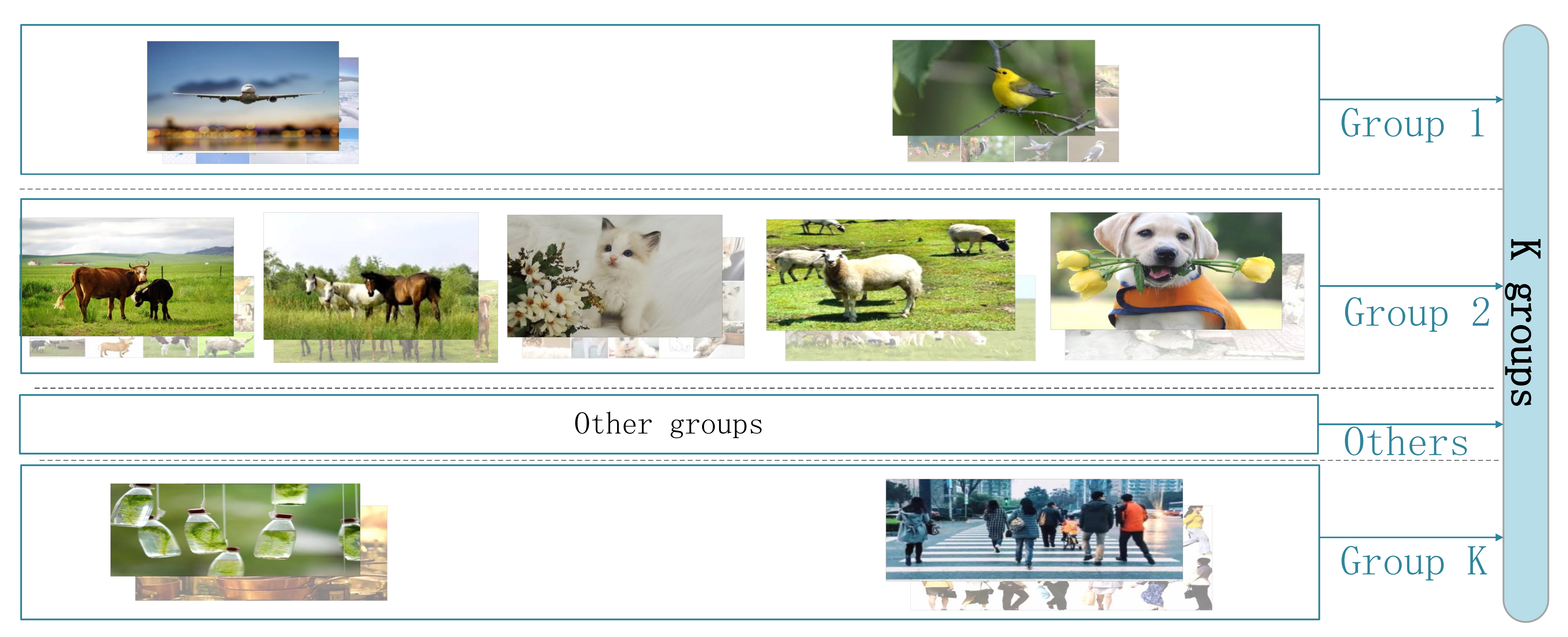}   
  \end{minipage}%
} 
\subfigure[Category-based grouping table on Pascal VOC dataset.]
{
  \begin{minipage}[c]{0.5\textwidth} 
  \center
    \begin{tabular}{|l|llllll|}
    \hline
      group&\multicolumn{6}{c|}{Class}\\
	\hline
    1& aero& bird&&&& \\   
    \hline
    2&cow&horse&cat&sheep&dog&\\   
    \hline
    3&sofa&chair&&&&\\  
    \hline
    4&tv&plant&table&&&\\
    \hline
    5&boat&bicycle&train&car&bus&mbike\\
    \hline
    6&bottle&person&&&&\\
    \hline
    \end{tabular}    
  \end{minipage}
}
  \caption{Overall scheme of category-based grouping mechanism. In (a), all categories are divided into $K$ groups. All categories in each row are similar in appearance and environment appeared. All categories in each row are a group. When ``bird'' is flying, it looks like ``aero''. ``cows, horses, cats, sheep and dogs'' have four legs and similar shape, and they often appear in similar environments. We regard them as a group. In this work, we experiment Pascal VOC dataset. As detailed in (b), all categories of Pascal VOC dataset are divided into 6 (i.e., K=6) groups by (a). The appearance and the appeared environment are very similar between categories within a group.}
 \label{figure1}
\end{figure*}
In order to solve those problems, based on \cite{kang2019few-shot}, we propose a new top-C classification loss (i.e., TCL-C) for few-shot learning to improve performance of few-shot detection. For YOLOv2 \cite{Redmon2017YOLO9000}, classification task is a binary classification task and ignores other results. Although the Cross-Entropy loss \cite{Rubinstein1999The} or Focal loss \cite{Lin2017Focal} can reduce the trust on the original label and increase the trust on the other labels to a certain extent. That cannot ensure that other categories-based features except for the true-label are bad for learning features, hurting detection performance because of few samples. Many researchers exploit the label smoothing and Cross-Entropy to alleviate the problem. However, they do not eliminate category-based irrelevant features and enhance similarity semantic features between categories. Smoothing label may increase irrelevant semantic features of other classes and hurt performance for few-shot learning. \textbf{Therefore, for classification, except for the true-label, we constraint the C-1 predictions with the highest classification scores. Because the C-1 false classes are the most likely to predict as the true-label, in this paper, we regard them as the most likely C-1 false classification predictions}, we can only set a simple constraint to enhance the semantic features relating with the true-label and suppress irrelevant semantic information. 

For few-shot detection, previous many methods\cite{wang2020frustratingly} \cite{yan2019meta}\cite{wang2019meta} \cite{hsieh2019one} \cite{hsieh2019one} \cite{fan2020few} \cite{zhu2021semantic} \cite{2021FSCE} exploit Faster R-CNN with FPN, multi-relation networks and ResNet with many other mechanisms as backbone to detect few-shot objects well. However, their networks are very complex, and they fail to consider strong bias problem (i.e., model can detect classes with sufficient training samples, and classes with the few samples is poor). In this paper, we propose a category-based grouping mechanism only by labels to allevaite strong bias problem. As shown in Figure \ref{figure1} (a), the left in the first row is the object "aero" and the last one in the same row is "bird", they are similar in appearance (i.e., visual appearance, shape, and limbs etc.). In most conditions, they often appear on the same environment. In this Figure, these are also very similar in appearance. And the scenes are also similar between objects in 1th, 2th, and 4th column, and between objects in 3th and 5th column. As seen in Figure \ref{figure1}, we split classes into disjoint groups. Therefore, this work proposes a category-based grouping mechanism to assist model learn meta-features better. Few methods without additional datasets or modules have found the characteristic, applying that into few-shot detection. Therefore, we alleviate the strong bias problem between classes and further improve few-shot detection APs by the category-based grouping. Based on the few-shot detection\cite{kang2019few-shot}, our contributions are as follows:
\begin{quote}
\begin{itemize}
\item We design a top-C classification loss (i.e.,TCL-C), which allows the true-label prediction and the most likely C-1 false classification predictions to improve performance on few-shot classes. 
\item Based similar appearance (i.e., visual appearance, shape, and limbs etc.) and environment in which objects often appear. We group categories into sub-groups with disjoint each other. Then, we construct a category-based grouping loss on meta-features grouped, which alleviates the strong bias problem and further improves detection APs.
\item We experiment different classification losses for few-shot detection on Pascal VOC, and all results show that our TCL-C performs better.
\item Combining the TCL-C with category-based grouping mechanism, for $k$-shot detection, $k=1, 2, 3,$ the detection APs achieve almost 20\%, 25\%, 30\%, respectively. Experimental results show that ours outperforms the state-of-the-art methods, and grouping is beneficial for concentration of detection APs between classes.
\end{itemize}
\end{quote}.
\section{Related Work}
\textbf{Classification Loss.} Different classification losses, such as the BCEwithLogits, the Cross-Entropy loss with SoftMax\cite{Rubinstein1999The}  \cite{liu2016large-margin} \cite{liu2017sphereface:}, are proposed. Most computer vision tasks use the Cross-Entropy to implement training. Then, \cite{Lin2017Focal} proposes a focal loss to alleviate the imbalance between the positive and negative samples. However,  many tasks based on YOLOv2 \cite{Redmon2017YOLO9000} just exploit a binary classification loss, resulting in the imbalance and ignoring the correlation about categories. In this paper, for few-shot classes, we assume that too much noise can hurt detection, and only the true-label may fail in learning relation with other categories. Therefore, we propose the TCL-C for classification task, which only focuses on the true-label and the most similar C-1 false classes. Compared with \cite{rahman2018polarity}, our TCL-C only exploits semantic information to promote performance.

\textbf{Meta-Learning.} Recently, different meta-learning algorithms have been proposed, including metric-based \cite{li2019revisiting} \cite{lifchitz2019dense} \cite{kim2019variational}, memory networks \cite{santoro2016meta-learning} \cite{oreshkin2018tadam:} \cite{mishra2017simple}, and optimization \cite{grant2018recasting} \cite{lee2018gradient-based} \cite{finn2017meta-learning} \cite{finn2017model-agnostic} \cite{kang2019few-shot}. The first type learns a metric based on few samples given and score a label of the target image according to similarity. The second is cross-task learning, and most memory networks widely are model-independent adaptation \cite{finn2017model-agnostic}. A model is learned on a variety of different tasks, making it possible to solve some new learning tasks with only few samples. Many researchers propose many variants \cite{nichol2018on, sun2019meta-transfer} \cite{antoniou2018how} \cite{rusu2019meta-learning}. The last one is a parameter prediction.\cite{kang2019few-shot} detects objects by Yolov2 \cite{Redmon2017YOLO9000}, and based on that, we further improve performance and alleviate many problems for few-shot detection.

\textbf{Few-shot Detection.} Previous most few-shot detection methods mainly focus on fine-tuning and metric learning, \cite{wang2020frustratingly}\cite{kang2019few-shot} and \cite{hsieh2019one} \cite{fan2020few} \cite{zhu2021semantic} \cite{2021FSCE}.\cite{wang2019meta} based on \cite{2017Faster} predicts parameters of category-specific by a weight prediction meta-model, but the category-agnostic is botained by base class samples. \cite{wang2020frustratingly} \cite{hsieh2019one} \cite{fan2020few} use FasterR R-CNN and ResNet-101 with FPN as backbone to detect objects. \cite{chen2018lstd:} transfers the basis domain to the novel. \cite{fan2020few} based on \cite{2017Faster} exploits attention-RPN and three relations to improve performance for few-shot detection. However, those methods fail to consider unequal detection APs and increase more parameters, resulting in training slower and poor performance on few-shot classes. \cite{2021FSCE} uses a contrastive branch measures the similarity between proposals, the method fails to the similarity between categories. \cite{zhu2021semantic} projects features into the category-based embedding space which is obtained by a large corpus of text, the cost of embedding space is much more. Therefore, based on \cite{kang2019few-shot}, we only split all categories into disjoint groups to improve detection performance without additional sub-modules(i.e., FPN, multi-relation modules, and metric function etc.), and captures the correlation between groups or categories from the category-based meta-features to reduce unequal detection.
\section{Our Approach}
\begin{figure*}
  \centering
  \includegraphics[width=2.0\columnwidth]{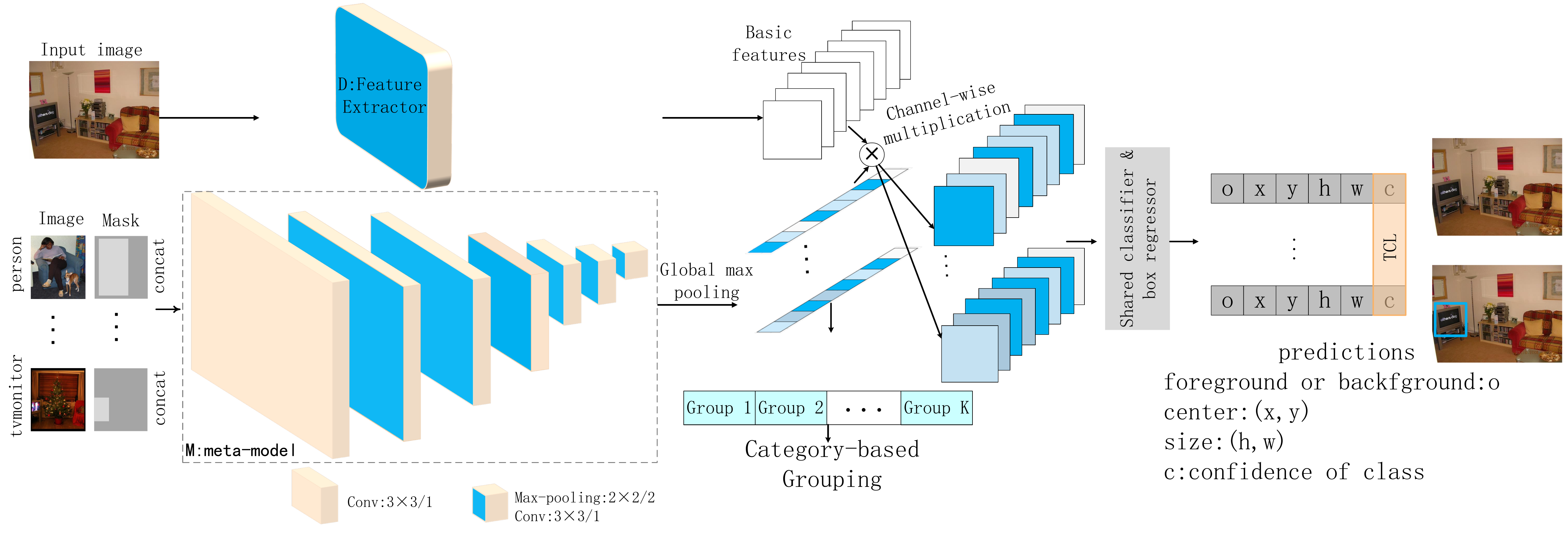}
  \caption{Overall structure of our method. The detection model consists of a feature extractor $D$ and a meta-model $M$.}
  \label{figure2}
\end{figure*}

As shown in Figure \ref{figure2}, we propose the TCL-C for classification and category-based grouping mechanism to help meta-model $M$ learn the related features between categories. The input of the meta-model $M$ is an image and a mask of only an object selected randomly. The value of the mask within object is 1, otherwise, it is 0. Every iteration, the M inputs the same number of samples as the number of categories. The meta-model M extracts meta-feature vectors about classes as the weight for reweighting features from feature extractor D, then, the classifier and detector complete classification and regression task. During training, we use the TCL-C and category-based grouping mechanism to train classification and help meta-model $M$, respectively. According to categories-based grouping mechanism, we split category-based meta-feature vectors into groups to learning better meta-features.
\subsection{Feature Reweighting for Detection}
Different categories may have a common and unique distribution.
As shown in Figure \ref{figure2}, based on YOLOv2 \cite{Redmon2017YOLO9000}, this method uses a meta-model $M$ to obtain meta-feature about categories for reweighting features. The meta-learning model takes each annotated sample ($I_i$, $B_i$), and for the category $i$, $i = 1, 2 ,..., N, N$ represents the number of categories. $I_i$ and $B_i$ represent the image and the selected mask of only an object on $ith$ class, obtaining category-based meta-feature by $M$. The $M$ learns to predict $N$ vectors $W$, $W=\{ w_1, w_2, w_3,...,w_N\}$, where $w_i$ represents the meta-feature vector of the $i-th$ category, $w_i = M (I_i, B_i)$, $M$ is the meta-model. Based on Darknet-19, the author builds a feature extractor $D$ which extracts basis features $F_j$ from the image $S_j$: $F_j = D(S_j)$. Then, for class $i$, the reweighted feature vector is obtained according to $w_i$ and $F_j$: $F_{j,i}$=$F_j\otimes$$w_i$. Finally, based on $F_{j,i}$, the author uses classifier and detector to classify and regress. See \cite{kang2019few-shot} for details
\subsection{TCL-C}
For few-shot detection, especially for meta-learning, we use the TCL-C to encourage model to train classification by the true-label prediction and the most likely C-1 false classification predictions, enhancing the semantic features relating with the true-label and suppressing irrelevant semantic information. As shown in Equation \ref{equation1}, our TCL-C makes the features tend to learn the true-label and controls effect of the most similar C-1 classes by ${\beta}^+$ and ${\beta_c}^-$, respectively, improving performance on novel (i.e.,few-shot) classes. Therefore, $\bm{{\beta}^+}$, $\bm{{\beta_c}^-}$ denote the expected classification score of true-label, and highest $k-th$ classes expected scores except for the true-label, respectively. Because of different similarities between categories,$\bm{{\beta_c}^-}$ is obtained through extensive experiments. Therefore, in this paper, our experiment about TCL-C is only 2(i.e.,C=2), as detailed in Equation\ref{equation1111}. Finally, $\eta$ and $\gamma$ affect the convergence rate. Detailed in Equation \ref{equation1} below.
\begin{equation}
\begin{split}
 L_{cls}&={L_{cls}}^{pos}+{L_{cls}}^{neg}\\
 {L_{cls}}^{pos}&=log(\eta+e^{\gamma({\beta}^+-P_t)})\\
 {L_{cls}}^{neg}=&log(\eta+e^{\gamma({F_t}^1-{\beta_1}^-)})+log(\eta+e^{\gamma({F_t}^2-{\beta_2}^-)})\\
&+\cdots+log(\eta+e^{\gamma({F_t}^c-{\beta_c}^-)})+\cdots+\\
&log(\eta+e^{\gamma({F_t}^C-{\beta_C}^-)}), 2\leq C\leq N \\
 \end{split}
\label{equation1}
\end{equation}
where ${L_{cls}}^{pos}$ and ${L_{cls}}^{neg}$ represent the loss functions for the true-label class and the most likely C-1 false classes, respectively. $P_t$ and $F_t$ represent the prediction score on the true-label class, and prediction scores of the most similar C-1 classes(i.e., the most likely C-1 false classes), respectively. $0\leq {\beta_k}^- \leq1$, ${\beta_k}^-$ is the expected threshold for the $kth$ (i.e., k=1, 2, ..., C) similar category. During analysis, as detailed in Equation \ref{equation1111}, we only experiment TCL-2. When C is bigger than 2 for TCL-C, how much influence other classes except for the true-label have on detection, we need to adjust the threshold value of each category dynamically to apply to the true-label, which requires extensive experiments, and we will experiment it in the future. Therefore, we urge the model to distinguish between the semantic features of the two most similar categories, improving the detection APs on few-shot classes. 
\begin{equation}
{L_{cls}}^{neg}=log(\eta+e^{\gamma(F_t-{\beta}^-)})\\
\label{equation1111}
\end{equation}
\subsection{Category-Based Grouping}
As detailed in Figure \ref{figure2}, meta-learning uses the correlation between categories for few-shot detection. As shown in Figure \ref{figure1}, our grouping mechanism focuses on appearance, followed by the environment, splitting all categories into $K$ groups which are disjoint with each other. We mainly analyze the mean and variance of the category-based meta-feature distribution from $M$. As for the principle (see Figure \ref{figure1}), as shown in Equation \ref{equation2},we propose a category-related loss about groups, the intra-group distance is smaller and the inter-group distance is larger. Our method encourages the variance of the mean value of the feature vector smaller for every group, making the semantic feature distribution more compact between categories within each group, and helps the feature distribution sparser between groups, improving detection APs and reducing the detection dispersion on the all categories.
\begin{equation}
L_{re-meta}=\sum\limits_{j=1}^Klog(\tau+{{L^j}_{group}})      
\label{equation2}
\end{equation}
As detailed in Equation \ref{equation2}, where ${{L^j}_{group}}$ is cost composed of within $kth$ group and between the $kth$ group and other groups, $\tau$ is set to 1.0. As shown in Equation \ref{groupj}.
\begin{equation}
\begin{split}
{{L^j}_{group}}=\frac{{W_{mean-std}}^j}{\epsilon+\frac{1}{{W_{mean-std}}^j}+\sum\limits_{k=j+1}^KL_{j,k}}\\
L_{j,k}=e^{{({W_{std}}^j-{W_{std}}^k)}^2} +e^{{({W_{mean}}^j-{W_{mean}}^k)}^2}
\end{split}
\label{groupj}
\end{equation}
where $L_{j,k}$ represents the meat-features distribution difference between the $jth$ group and the $kth$ group, we expect the value to be as big as possible. $\bm{{W_{mean-std}}^j}$ represent the dispersion of concentration of the meta-feature space between categories within the $jth$ group, and we expect it to be as small as possible (i.e., distribution between categories within gruops is more compact). $\bm{{W_{std}}^j}$ and $\bm{{W_{mean}}^j}$ represents the dispersion metric and the concentration metric of meta-features between categories within the $jth$ group , respectively. According to Equation \ref{equation2}, we expect the distribution of different categories within the groups is more compact, and the different groups are far farther from each other (i.e., distribution is obvious between groups). $\bm{{W_{mean-std}}^j}$ is smaller, then, $\bm{{W_{std}}^j-{W_{std}}^k}$ and $\bm{{W_{mean}}^j-{W_{mean}}^k}$ are bigger. Every theory is detailed below.
\begin{equation}
{W_{mean-std}}^j=\sqrt{ \frac{1}{||C_j||}  \sum\limits_{m=1}^{||C_j||}{(   {u_m}^j - u^j )}^2 }
\label{equation3}
\end{equation}
where we expect the value to be smaller. $C_j$ is the $jth$ group, and $||C_j||$ represents the number of classes within the $jth$ group. $u^j$ is the mean value of all features for the $j$th group, and ${u_m}^j$ is the mean value of meta-features for the $mth$ class of the $jth$ group. 
\begin{equation}
\label{equation41}
{W_{std}}^j=\left\{
\begin{aligned}
\sqrt{ \frac{1}{||C_j||}  \sum\limits_{m=1}^{||C_j||}{(   {\delta_m}^j - \delta^j )}^2 } & , &||C_j||>1. \\
{\delta_m}^j& , &||C_j||=1.
\end{aligned}
\right.
\end{equation}
\begin{equation}
\label{equation42}
{W_{mean}}^j=\left\{
\begin{aligned}
\frac{1}{||C_j||}  \sum\limits_{m=1}^{||C_j||}   {u_m}^j & , &||C_j||>1. \\
{u_m}^j& , &||C_j||=1.
\end{aligned}
\right.
\end{equation}
\begin{equation}
\label{equation5}
u_i=\frac{1}{|F|} \sum\limits_{f=1}^{|F|}  {x_f}^i , \quad \delta_i=\sqrt{  \frac{1}{|F|}  \sum\limits_{f=1}^{|F|}  {(  {x_f}^i-   u_i )}^2 }
\end{equation}
where $x$ denotes that the category-related $|F|$-dimension meta-feature vectors, $|F| = 1024$, $i=1,2,3,...,N$. $u_i$ and $\delta_i$ represent the mean value and variance of meta-feature for the $ith$ category, respectively. Within the $jth$ group, ${\delta_m}^j$ and $\delta^j$ are the variance value of the $mth$ class and the variance value of all meta-features, respectively. As shown in Figure \ref{figure1}, all 20 categories are divided into 6 groups, $K = 6$, $C_j\in\{C_1, C_2, ..., C_6\}$. Because of the correlation loss of each group, the value of the $log$ function is less than 0. Therefore, the parameter $\tau$ is used to ensure that the loss is a positive value, and the parameter must be greater than or equal to 1. In terms, the method alleviates the phenomenon which the performance on different categories varies greatly for few-shot detection. See Appendix A for details.
\subsection{Loss Details}
\textbf{Category-Based Grouping.} Considering that different categories in different environments have the similar appearance and different categories are in the similar environment, in order to reduce the setting, we mainly focus on the appearance similarity, followed by the environment, and we set classes with the similar appearance and scenes appeared as a group. As shown in Figure \ref{figure1} (b), we divide the Pascal VOC with 20 categories into 6 groups, namely $K = 6$. As shown in the Equation \ref{equation2}, we set the parameter $\tau$, $\epsilon$ to 1 and 0.00005 respectively.

\textbf{Loss Functions.} In order to train the meta-model and ensure that the shared features ( i.e., within a group) are more compact between the meta-features of the similar semantic objects, we jointly train classification, category-based grouping, and regression, as shown in Equation \ref{equation6}. Compared with state-of-the-art classification methods, our TCL-2 method is more suitable for few-shot detection.
\begin{equation}
\begin{split}
L_{loc}&=L_{loc}(x)+L_{loc}(y)+L_{loc}(w)+L_{loc}(h)\\
L&=\alpha L_{cls}+\omega L_{re-meta}+\lambda L_{loc}
\end{split}
\label{equation6}
\end{equation}
where $ L_{re-meta}$ denotes the category-based grouping loss. $ L_{loc}$ includes the center location loss $ L_{loc}(x)$, $ L_{loc}(y)$ and scale loss $ L_{loc}(w)$, $ L_{loc}(h)$.  In this experiment, the classification, similarity, and regression balance parameters, $\alpha$, $\omega$ and $\lambda$,  are set to 1, 6, and 1, respectively.
\section{Experiments and Results}
This experiment consists of the base training and few-shot fine-tuning. As shown in Figure \ref{figure2}, the output of the meta-model $M$ is related to the number of categories, and each category-based meta-feature vector is represented as a 1024-dimension vector. We experiment with different classification losses, BCEwithLogits, Focal \cite{Lin2017Focal}, Cross-Entropy \cite{Rubinstein1999The} and the TCL-2, combining with the Category-based Grouping, respectively. Methods which combine all classification losses and our proposed category-based grouping are regarded as Re-BCEwithLogits, Re-Focal, Re-Cross-Entropy, and Ours (i.e.,Re-TCL-2), respectively. Detail as follows.
\subsection{DataSets and Setting}
The Pascal VOC DataSet contains 20 categories, we randomly select 5 categories as novel(i.e., few-shot) categories with few samples (i.e.,k-shot, $k=1,2,3,5$) for fine-tuning, and the remaining 15 categories as base classes with sufficient samples for the base model. The 20 categories are randomly divided into 6 novel parts, and we experiment 3 parts obtained as the novel classes for fine-tuning $k$-shot, $k=1,2,3,5$. Our setting is the same with \cite{kang2019few-shot}. During base classes training, only samples with 15 categories are trained, and the remaining regarded as novel classes with 5 categories are fine-tuned, and each novel class has only $k$-shot, named 5-way $k$-shot. For meta-model, when there are multiple objects in an image, only an object mask is randomly selected corresponding to the classes. All models are trained by 4 GPUs with 64 batch sizes, and we train for 80,000 iterations for base model. In our work, we use test sets of VOC2007 as our test sets, and training/validation sets of VOC2007 and VOC2012 as our training sets. We use SGD with momentum 0.9, and L2 weight-decay 0.0005 for detector and meta-model.
\subsection{Ablation Studies}
Our experiments are mainly for 5-way $k$-shot. We analyze the detection performance on Pascal VOC by the category-based meta-feature grouping and different classification losses. The details are as follows.
\begin{table*}
\small
  \centering
     \begin{center}
  \begin{tabular}{|l|llll|llll|llll|}
    \hline
    &\multicolumn{4}{c|}{Novel Set1} & \multicolumn{4}{c|}{Novel Set2} &\multicolumn{4}{c|}{Novel Set3}\\
     \hline
     \centering
    Method/shot& 1& 2& 3& 5 &  1& 2& 3& 5 &  1& 2& 3& 5 \\
     \hline
    \centering
    BCEwithLogits&16.42&18.51&27.41&36.07&13.59&14.71&26.3&35.2&15.1&15.62&26.14&31.6\\   
   Re-BCEwithLogits&13.26&17.46&24.31&33.76&\textbf{18.29}&19.71&26.99&35.3&11.0&13.0&20.74&31.95\\
   \hline
    Focal&16.27&21.63&27.91&\textbf{37.43}&10.39&15.23&18.36&34.09&9.6&8.87&20.16&27.54\\
    Re-Focal&18.22&20.05&20.45&36.15&14.16&15.88&23.13&27.2&7.3&8.67&16.35&28.6\\
    \hline
    Cross-Entropy&15.37&19.11&23.11&35.18&16.04&19.2&25.46&35.84&12.19&15.3&20.31&31.91\\
    Re-Cross-Entropy&18.55&21.02&22.25&36.5&15.15&20.81&26.07&33.45&13.07&14.53&23.93&35.58\\ 
    \hline   
    TCL-2 &19.15&21.23&28.64&36.94&17.56&22.25&25.57&38.45&12.27&17.33&\textbf{30.81}&35.08\\
    Ours&\textbf{20.08}&\textbf{26.75}&\textbf{29.76}&36.28&18.07&\textbf{24.66}&\textbf{30.94}&\textbf{39.04}&\textbf{19.42}&\textbf{17.43}&23.24&\textbf{37.66}\\
    \hline
  \end{tabular}
 \end{center}
  \label{sum_result}
   \caption{The results of detection APs(\%) on novel classes. For few-shot detection on Pascal VOC, our method significantly outperforms others.}
\end{table*}
\begin{figure}[htb] 
\subfigure[ Curves of the APs(\%) normalized for all methods.]
{
  \begin{minipage}[c]{0.47\textwidth} 
    \centering 
    \includegraphics[width=0.93\textwidth]{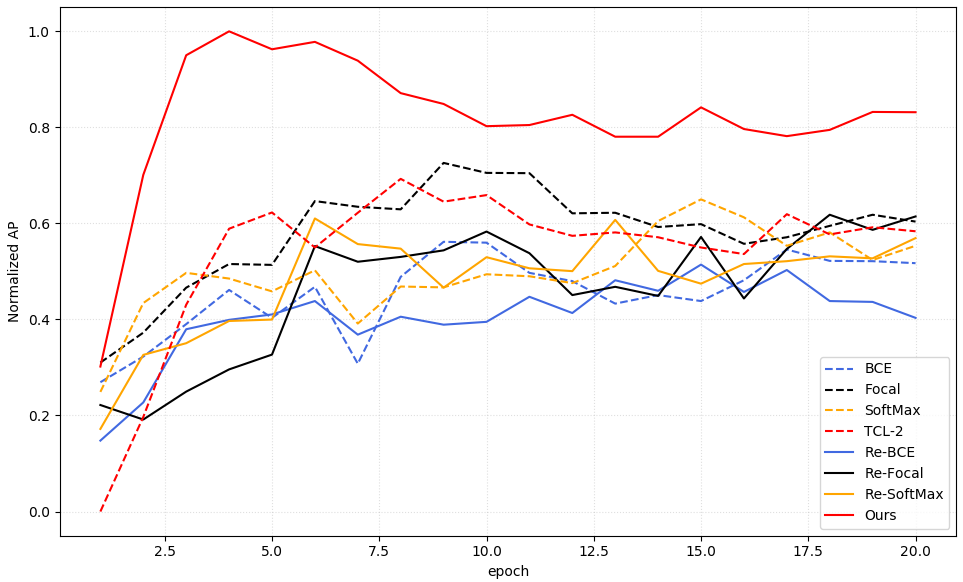}       
  \end{minipage}%
}
\subfigure[ Curves of the APs(\%) normalized for different $\beta^-$. ]
{
  \begin{minipage}[c]{0.5\textwidth} 
    \centering 
    \includegraphics[width=0.95\textwidth]{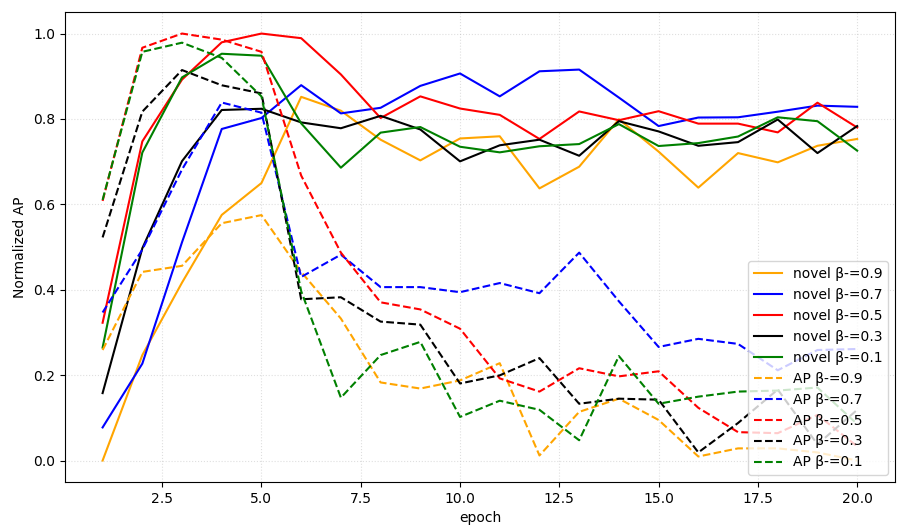}       
  \end{minipage}%
}
  \caption{ For Pascal VOC, in (a), for novel set1 2-shot, the curve shows all epochs fine-tuning results of the detection APs (\%) on the novel classes. Our method obviously outperforms others. In (b), Results with solid line are normalized APs on the novel classes, and results with the dashed line are the detection normalized APs on all categories. For TCL-2, when the $\beta^-$ is set to 0.5, our method is better for fine-tuning novel1 2-shot.}
    \label{beta}
\end{figure}
\subsubsection{The Importance of the TCL} 
 \textbf{Impact of $\beta^-$.} As shown in Equation \ref{equation1}, $\beta^-$ is essential for our TCL-2. The threshold of true-label $\beta^+$ is 1.0, the threshold of the most likely false classification prediction $\beta^-$ is set to 0.5. $\beta^-$ cannot be too large or small. If it is too large, making meta-model drive the other classes towards the true-label,and that causes semantically similar categories to become ambiguous and inconducive to classification. Otherwise, it makes the model trust only the true-label,and that make farther gap between similar categories and fail to make full use of semantic to assist each other. As illustrated in Figure \ref{beta}(b), when the $\beta^-$ is greater than 0.5, the APs distribution on the novel classes(see the solid line) is consistent, and best APs on all categories(see the dashed line) is lower than our setting, $\beta^-=0.5$. When $\beta^-$ is less than 0.5, the APs on the novel classes tend to be smooth because the semantics between the true-label and the false most likely classification results are clearly separated, making model trust the true-label most and failing to weaken the assistance of false true result. Therefore, when the $\beta^-$ is set to 0.5, our method can exploit similar semantics distribution between different categories to improve the performance on novel classes better. 

\textbf{Comparsion with  the state-of-the-art losses.} As shown in Table 1, 
compared with the state-of-the-art BCEwithLogits, Focal \cite{Lin2017Focal}, and Cross-Entropy \cite{Rubinstein1999The}, our TCL-2 can improve the few-shot detection performance. For novel set 1, the 1-shot detection APs of TCL-2 is 2.73\%, 2.88\%, and 3.78\% better than the other classification losses, respectively. The TCL-2 is better than the other classification methods by 1.23\%, 0.73\%, and 5.53\% for 3-shot, respectively.
 TCL-2 has another advantage (i.e.,alleviating the strong bias problem). As shown in Table 2, for example, our TCL-2 for dispersion of detection APs is 2.04\%, 4.0\%, and 4.15\% better than the BCEwithLogits, Focal and Cross-Entropy on novel set 1, respectively.
\begin{table*}
\small
  \centering
   \resizebox{1.0\textwidth}{!}{
  \centering
  \begin{tabular}{|l|ll|ll|ll|ll|}
    \hline    
    Shot/Method& BCEwithLogits& Re-BCEwithLogit& Focal& Re-Focal&Cross-Entropy&Re-Cross-Entropy&TCL-2&Ours  \\
    \hline
    \centering
    1&55.04&58.58&57.0&54.77&57.15&53.86&53.0&\textbf{52.63}\\
    \hline
    2&52.84&54.46&57.0&51.99&53.36&50.63&50.93&\textbf{46.45}\\
    \hline
    3&45.81&48.28&45.67&51.22&49.63&48.6&45.46&\textbf{44.37}\\
    \hline
    5&41.1&42.07&\textbf{40.74}&32.57&41.95&40.64&41.82&41.36\\
\hline
  \end{tabular}
  }    
    \label{divergence_fine}
    \caption{Dispersion of the detection APs(\%) on all categories. For novel set1, our method obviously alleviates the strong bias, reducing dispersion of detection performance.}
\end{table*}
\subsubsection{Analysis of the Category-based Grouping}
\textbf{Impact on every strategy.} We design the scheme in Equation \ref{equations} by category-based grouping mechanism (i.e., Figure \ref{figure1} and Equation \ref{equation2}). Then, the experiment analyzes every component (i.e., $q_j$, $Q_j$ and $U_j$).
\begin{equation}
L_{re-meta}=\sum\limits_{j=1}^Klog(\tau+\frac{q_j}{\epsilon+Q_j+\sum\limits_{k=j+1}^KU_j}).
\label{equations}
\end{equation}
Except for the best strategy (Equation \ref{equation2} and Equation \ref{groupj}), we experiment three other strategies, details as follows:
\begin{equation}
q_j=1, Q_j=0, U_j=e^{{({W_{std}}^j-{W_{std}}^k)}^2}
\label{equation10}
\end{equation}
where related grouping is only related to dispersion of meta-feature distribution between groups, the method fails to take into account the similarity of meta-feature distribution between groups. Therefore, we optimize the component $U_j$, as detailed below:
\begin{equation}
\begin{split}
q_j&=1, Q_j=0\\
U_j&=e^{{({W_{std}}^j-{W_{std}}^k)}^2} +e^{{({W_{mean}}^j-{W_{mean}}^k)}^2}
\end{split}
\label{equation11}
\end{equation}
 $U_j$ can learn features between groups. However, the method fails to learn meta-feature distribute within a group.
Then,our best category-based grouping metric (Equation \ref{equation2}) makes the distribution of meta-features more compact within a group and the difference between groups more obvious.
In the other hand, if the category-based grouping is only attribute to the disperse and similarity of meta-feature between groups, grouping mechanism cannot learn the difference of meta-features between categories within a group, failing to distinguish between different categories of meta-feature distribution within a group, detail as follow:
\begin{equation}
\begin{split}
q_j&={W_{mean-std}}^j, Q_j=\frac{1}{{W_{mean-std}}^j}.\\
U_j&=e^{{({W_{std}}^j-{W_{std}}^k)}^2}.
\end{split}
\label{equation12}
\end{equation}
As shown in Table 3 
, for every strategy ( Equation \ref{equation2}, \ref{equations}, \ref{equation10},\ref{equation11},and \ref{equation12}), we experiment different Re-TCL methods which combine our TCL-2 with different category-based grouping methods. Ours combining TCL-2 with the strategy (in Equation \ref{equation2}) is better for few-shot detection.
 \begin{table*}
\small
  \centering
  \begin{tabular}{|ll|llllll|l|l|}
    \hline
    \centering
    & &\multicolumn{6}{c|}{Novel Set 1} &\multicolumn{2}{c|}{APs}\\
    \hline
    \centering
     Shot&Method&boat&cat&mbike&sheep&sofa&mean&base AP& AP\\
     \hline
     \centering    
   \multirow{4}{*}{1}& Re-TCL-2 (Equation \ref{equations} and \ref{equation10})&9.39&\textbf{37.6}&28.63&17.93&12.84&\textbf{21.27}&65.72&54.61\\
   &Re-TCL-2 (Equation \ref{equations} and \ref{equation11})&4.55&32.78&29.89&18.28&10.87&19.27&\textbf{66.51}&\textbf{54.7}\\
   & Re-TCL-2 (Equation(\ref{equations} and \ref{equation12})&9.09&34.75&21.63&17.11&\textbf{19.67}&20.45&65.14&53.97\\
   &Ours \ref{equation2}&\textbf{9.53}&33.58&\textbf{32.28}&\textbf{19.66}&5.34&20.08&65.44&54.1\\
  \hline
     \multirow{4}{*}{2}& Re-TCL-2 (Equation \ref{equations} and \ref{equation10})&7.22&41.24&20.34&32.36&13.66&22.96&64.83&\textbf{55.61}\\
   &Re-TCL-2 (Equation \ref{equations} and \ref{equation11})&5.22&39.38&\textbf{33.79}&33.46&11.9&24.75&65.05&54.97\\
   &Re-TCL-2 (Equation \ref{equations} and \ref{equation12})&2.19&\textbf{45.05}&25.01&27.84&17.8&23.56&64.83&54.51\\
   &Our \ref{equation2}&\textbf{10.61}&35.11&33.75&\textbf{35.89}&\textbf{18.38}&\textbf{26.75}&\textbf{65.18}&55.58\\
  \hline
     \multirow{4}{*}{3}& Re-TCL-2 (Equation \ref{equations} and \ref{equation10})&6.32&\textbf{47.77}&22.45&27.92&29.99&26.89&65.55&55.89\\
   &Re-TCL-2 (Equation \ref{equations} and \ref{equation11})&\textbf{10.46}&47.35&27.08&26.12&\textbf{37.72}&29.75&65.11&56.27\\
   &Re-TCL-2 (Equation \ref{equations} and \ref{equation12})&10.29&39.55&18.76&28.67&33.84&26.22&65.18&55.44\\
   &Ours \ref{equation2} &10.29&46.05&\textbf{28.11}&\textbf{29.81}&34.52&\textbf{29.76}&\textbf{65.64}&\textbf{56.67}\\   
\hline
  \end{tabular} 
  \label{compare_related_methods}
  \caption{The detection APs(\%) on Pascal VOC by combining our TCL-2 with different category-based grouping methods. For $k$-shot detection, $k=1,2,3$, ours is better than others for novel set1. This table illustrates APs (i.e., every novel class, the mean APs on the novel classes, the mean APs on the base classes, and the mean APs on the all categories).}
\end{table*} 

\textbf{Impact of category-based grouping mechanism.} Without additional datasets, as detailed in Equation\ref{equation2}, we mainly focus on the similar appearance between categories, followed by similar scenes, exploiting the relationships to promote performance. For Equation \ref{equation2}, we analyze category-based grouping, and compare with every ablation. As shown in Table 1, 
compared with only classification losses, splitting 20 categories into 6 disjoint groups can improve performance for few-shot detection.

\textbf{Impact on dispersion.} As shown in Table 2, 
Re-BCEwithLogits, Re-Focal, and Re-Cross-Entropy is compared with the BCEwithLogits, Focal, and Cross-Entropy, respectively. We find that the better meta-feature distribution between categories can alleviate the unbalanced performance on all categories. Especially for novel set1 2-shot, the dispersion of Re-Focal and Re-Cross-Entropy are reduced by 5.01\% and 2.73\%, respectively. Therefore, our category-based grouping mechanism can help distribution between similar semantic classes more compact and exploit the correlation between categories better, and alleviate the strong bias problem.

\begin{figure*}
  \centering
  \includegraphics[width=2.0\columnwidth]{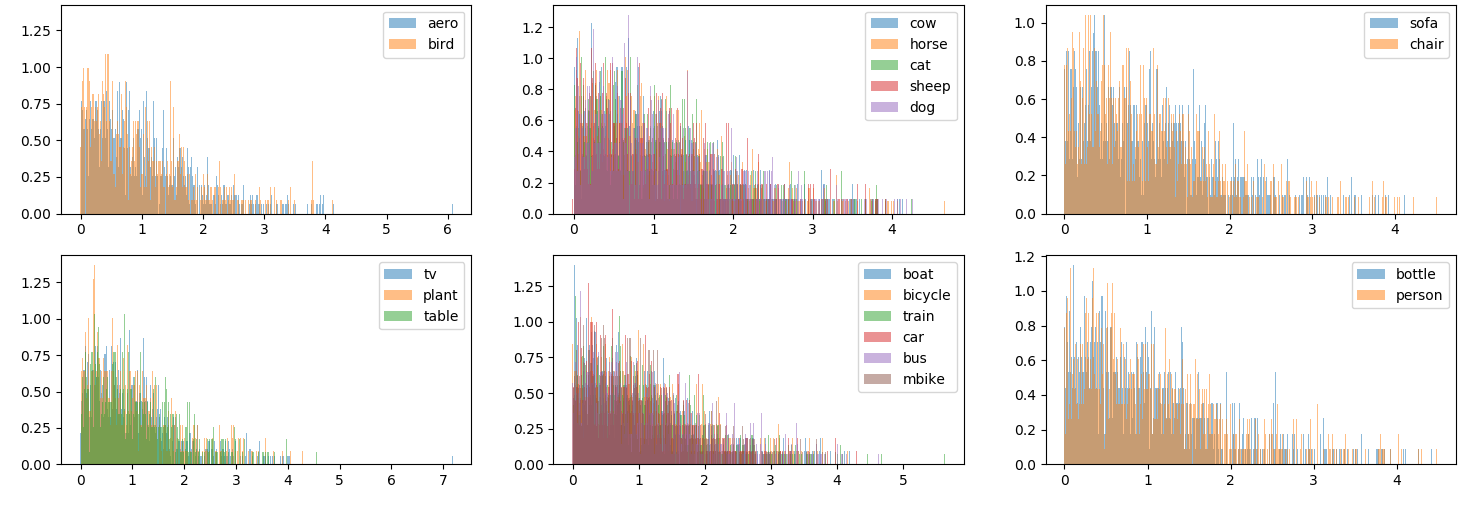}
  \caption{Histogram of meta-features distribution. The distribution of meta-feature vectors of 20 categories. In a sub-figure, the feature vectors of each group are represented, and different colors represent different categories within a group. Every meta-feature in a subgraph is very similar, and the distribution difference between subgraphs is obvious. Our method can extract shared meta-features within a group better.}
  \label{figure4}
\end{figure*}

As shown in Figure \ref{figure4}, for a subgraph, each category-based meta-feature is represented by different color histograms, and each subgraph is represented as a group with categories. We find that the meta-features distribution is very similar within a group, and the difference between groups is obvious.
\subsection{Visualization and Results}
Our TCL-$C$ and category-based grouping can improve detection performance on few-shot classes, as seen in Appendix B. First, as detailed in Equation \ref{equation1} and Figure \ref{beta} (a), our TCL-2 performs better for similar semantics to improve the detection APs on few-shot classes. 

Then, according to the similar appearance and environment which different categories appear, as detailed in Figure \ref{figure1} and Equation \ref{equation2}, we split all categories into $K$ groups which is disjoint each other. The distribution of meta-features is compact between categories within a group, the distribution between groups is far away from each other. That category-based grouping exploits similar distribution of category-based meta-features and reduces the detection dispersion on all categories. As can be seen from Figure \ref{beta} (a), Figure \ref{figure4} and Table 1, 
the category-based grouping helps meta-model extract the shared meta-features between categories and ours improves the detection APs by similar semantics between categories. As shown in Table 2, ours reduces the dispersion of the detection APs on all classes. 

Combining TCL-2 with category-based grouping is more beneficial for few-shot detection. As shown in Table 1, 
for $k$-shot detection, $k = 1,2,3, $ours is better, and the detection APs are close to 20.0\%, 25\%, 30\%, respectively. For novel set3, the novel classes are " aero, bottle, cow, horse, sofa" and the remaining is the base classes. Although there is no novel class associated with the base categories, our method is 4.32\%, 9.82\% and 6.35\% better than the BCEwithLogits, the Focal and Re-Cross-Entropy for novel set3 1-shot, respectively.  
\section{Conclusions}
For few-shot detection, we present a TCL-$C$ for exploiting the true-label and the most similar C-1 classes to improve the detection performance on few-shot classes, and a category-based grouping method for helping meta-model extract category-related meta-features better, alleviating the strong bias problem and further improving performance. Based on similar appearance or the environment they often appeared, this paper splits categories into disjoint groups manually. This method helps the meta-model extract meta-feature vectors, making distribution of meta-features within a group more compact and difference of meta-features more obvious between groups. For 1-shot, 2-shot, and 3-shot detection, our method obtains all detection APs of almost 20\%, 25\%, and 30\%. In the future, rather than group manually, we will combine categories-embedded features and unsupervised clustering to group more categories dynamically to improve performance for few-shot detection.

{\small
\bibliographystyle{ieee_fullname}
\bibliography{egpaper_final}

\begin{thebibliography}{10}\itemsep=-1pt

\bibitem{antoniou2018how}
Antreas Antoniou, Harrison Edwards, and Amos Storkey.
\newblock How to train your maml.
\newblock {\em arXiv: Learning}, 2018.

\bibitem{bansal2018zero-shot}
Ankan Bansal, Karan Sikka, Gaurav Sharma, Rama Chellappa, and Ajay Divakaran.
\newblock Zero-shot object detection.
\newblock pages 397--414, 2018.

\bibitem{chen2018lstd:}
Hao Chen, Yali Wang, Guoyou Wang, and Yu Qiao.
\newblock Lstd: A low-shot transfer detector for object detection.
\newblock pages 2836--2843, 2018.

\bibitem{fan2020few}
Qi Fan, Wei Zhuo, Chi-Keung Tang, and Yu-Wing Tai.
\newblock Few-shot object detection with attention-rpn and multi-relation
  detector.
\newblock In {\em Proceedings of the IEEE\/CVF Conference on Computer Vision
  and Pattern Recognition}, pages 4013--4022, 2020.

\bibitem{finn2017model-agnostic}
Chelsea Finn, Pieter Abbeel, and Sergey Levine.
\newblock Model-agnostic meta-learning for fast adaptation of deep networks.
\newblock pages 1126--1135, 2017.

\bibitem{finn2017meta-learning}
Chelsea Finn and Sergey Levine.
\newblock Meta-learning and universality: Deep representations and gradient
  descent can approximate any learning algorithm.
\newblock {\em arXiv: Learning}, 2017.

\bibitem{grant2018recasting}
Erin Grant, Chelsea Finn, Sergey Levine, Trevor Darrell, and Thomas~L
  Griffiths.
\newblock Recasting gradient-based meta-learning as hierarchical bayes.
\newblock {\em arXiv: Learning}, 2018.

\bibitem{hsieh2019one}
Ting-I Hsieh, Yi-Chen Lo, Hwann-Tzong Chen, and Tyng-Luh Liu.
\newblock One-shot object detection with co-attention and co-excitation.
\newblock {\em arXiv preprint arXiv:1911.12529}, 2019.

\bibitem{ShahTask}
Muhammad~Abdullah Jamal and Guo-Jun Qi.
\newblock Task agnostic meta-learning for few-shot learning.
\newblock In {\em Proceedings of the IEEE/CVF Conference on Computer Vision and
  Pattern Recognition}, pages 11719--11727, 2019.

\bibitem{kang2019few-shot}
Bingyi Kang, Zhuang Liu, Xin Wang, Fisher Yu, Jiashi Feng, and Trevor Darrell.
\newblock Few-shot object detection via feature reweighting.
\newblock pages 8420--8429, 2019.

\bibitem{kim2019variational}
Junsik Kim, Taehyun Oh, Seokju Lee, Fei Pan, and In~So Kweon.
\newblock Variational prototyping-encoder: One-shot learning with prototypical
  images.
\newblock {\em arXiv: Computer Vision and Pattern Recognition}, 2019.

\bibitem{kong2019foveabox}
Tao Kong, Fuchun Sun, Huaping Liu, Yuning Jiang, and Jianbo Shi.
\newblock Foveabox: Beyond anchor-based object detector.
\newblock {\em arXiv preprint arXiv:1904.03797}, 2019.

\bibitem{lee2018gradient-based}
Yoonho Lee and Seungjin Choi.
\newblock Gradient-based meta-learning with learned layerwise metric and
  subspace.
\newblock pages 2927--2936, 2018.

\bibitem{li2019revisiting}
Wenbin Li, Lei Wang, Jinglin Xu, Jing Huo, Yang Gao, and Jiebo Luo.
\newblock Revisiting local descriptor based image-to-class measure for few-shot
  learning.
\newblock pages 7260--7268, 2019.

\bibitem{lifchitz2019dense}
Yann Lifchitz, Yannis Avrithis, Sylvaine Picard, and Andrei Bursuc.
\newblock Dense classification and implanting for few-shot learning.
\newblock pages 9258--9267, 2019.

\bibitem{Lin2017Focal}
Tsung~Yi Lin, Priyal Goyal, Ross Girshick, Kaiming He, and Piotr Dollar.
\newblock Focal loss for dense object detection.
\newblock {\em IEEE Transactions on Pattern Analysis \& Machine Intelligence},
  PP(99):2999--3007, 2017.

\bibitem{liu2017sphereface:}
Weiyang Liu, Yandong Wen, Zhiding Yu, Ming Li, Bhiksha Raj, and Le Song.
\newblock Sphereface: Deep hypersphere embedding for face recognition.
\newblock pages 6738--6746, 2017.

\bibitem{liu2016large-margin}
Weiyang Liu, Yandong Wen, Zhiding Yu, and Meng Yang.
\newblock Large-margin softmax loss for convolutional neural networks.
\newblock {\em arXiv: Machine Learning}, 2016.

\bibitem{mishra2017simple}
Nikhil Mishra, Mostafa Rohaninejad, Xi Chen, and Pieter Abbeel.
\newblock A simple neural attentive meta-learner.
\newblock {\em arXiv preprint arXiv:1707.03141}, 2017.

\bibitem{HebertWatch}
Ishan Misra, Abhinav Shrivastava, and Martial Hebert.
\newblock Watch and learn: Semi-supervised learning for object detectors from
  video.
\newblock In {\em Proceedings of the IEEE Conference on Computer Vision and
  Pattern Recognition}, pages 3593--3602, 2015.

\bibitem{XieDual}
Jian Ni, Shanghang Zhang, and Haiyong Xie.
\newblock Dual adversarial semantics-consistent network for generalized
  zero-shot learning.
\newblock {\em arXiv preprint arXiv:1907.05570}, 2019.

\bibitem{nichol2018on}
Alex Nichol, Joshua Achiam, and John Schulman.
\newblock On first-order meta-learning algorithms.
\newblock {\em arXiv: Learning}, 2018.

\bibitem{oreshkin2018tadam:}
Boris~N Oreshkin, Pau~Rodriguez Lopez, and Alexandre Lacoste.
\newblock Tadam: Task dependent adaptive metric for improved few-shot learning.
\newblock pages 721--731, 2018.

\bibitem{rahman2018polarity}
Shafin Rahman, Salman Khan, and Nick Barnes.
\newblock Polarity loss for zero-shot object detection.
\newblock {\em arXiv preprint arXiv:1811.08982}, 2018.

\bibitem{PorikliZero}
Shafin Rahman, Salman Khan, and Fatih Porikli.
\newblock Zero-shot object detection: Learning to simultaneously recognize and
  localize novel concepts.
\newblock In {\em Asian Conference on Computer Vision}, pages 547--563.
  Springer, 2018.

\bibitem{Redmon2015You}
Joseph Redmon, Santosh Divvala, Ross Girshick, and Ali Farhadi.
\newblock You only look once: Unified, real-time object detection.
\newblock 2015.

\bibitem{Redmon2017YOLO9000}
Joseph Redmon and Ali Farhadi.
\newblock Yolo9000: Better, faster, stronger.
\newblock In {\em IEEE Conference on Computer Vision \& Pattern Recognition},
  2017.

\bibitem{ZemelIncremental}
Mengye Ren, Renjie Liao, Ethan Fetaya, and Richard~S Zemel.
\newblock Incremental few-shot learning with attention attractor networks.
\newblock {\em arXiv preprint arXiv:1810.07218}, 2018.

\bibitem{2017Faster}
Shaoqing Ren, Kaiming He, Ross Girshick, and Jian Sun.
\newblock Faster r-cnn: Towards real-time object detection with region proposal
  networks.
\newblock {\em IEEE Transactions on Pattern Analysis \& Machine Intelligence},
  39(6):1137--1149, 2017.

\bibitem{Rubinstein1999The}
Reuven Rubinstein.
\newblock The cross-entropy method for combinatorial and continuous
  optimization.
\newblock {\em Methodology \& Computing in Applied Probability}, 2(2):127--190,
  1999.

\bibitem{rusu2019meta-learning}
Andrei~A Rusu, Dushyant Rao, Jakub Sygnowski, Oriol Vinyals, Razvan Pascanu,
  Simon Osindero, and Raia Hadsell.
\newblock Meta-learning with latent embedding optimization.
\newblock 2019.

\bibitem{santoro2016meta-learning}
Adam Santoro, Sergey Bartunov, Matthew Botvinick, Daan Wierstra, and Timothy
  Lillicrap.
\newblock Meta-learning with memory-augmented neural networks.
\newblock pages 1842--1850, 2016.

\bibitem{2021FSCE}
Bo Sun, Banghuai Li, Shengcai Cai, Ye Yuan, and Chi Zhang.
\newblock Fsce: Few-shot object detection via contrastive proposal encoding.
\newblock 2021.

\bibitem{sun2019meta-transfer}
Qianru Sun, Yaoyao Liu, Tatseng Chua, and Bernt Schiele.
\newblock Meta-transfer learning for few-shot learning.
\newblock pages 403--412, 2019.

\bibitem{tian2019fcos}
Zhi Tian, Chunhua Shen, Hao Chen, and Tong He.
\newblock {FCOS}: Fully convolutional one-stage object detection.
\newblock In {\em Proc. Int. Conf. Computer Vision (ICCV)}, 2019.

\bibitem{wang2020frustratingly}
Xin Wang, Thomas~E Huang, Trevor Darrell, Joseph~E Gonzalez, and Fisher Yu.
\newblock Frustratingly simple few-shot object detection.
\newblock {\em arXiv preprint arXiv:2003.06957}, 2020.

\bibitem{wang2019meta}
Yu-Xiong Wang, Deva Ramanan, and Martial Hebert.
\newblock Meta-learning to detect rare objects.
\newblock In {\em Proceedings of the IEEE/CVF International Conference on
  Computer Vision}, pages 9925--9934, 2019.

\bibitem{wertheimer2019few-shot}
Davis Wertheimer and Bharath Hariharan.
\newblock Few-shot learning with localization in realistic settings.
\newblock pages 6558--6567, 2019.

\bibitem{PinheiroAdaptive}
Chen Xing, Negar Rostamzadeh, Boris~N Oreshkin, and Pedro~O Pinheiro.
\newblock Adaptive cross-modal few-shot learning.
\newblock {\em arXiv preprint arXiv:1902.07104}, 2019.

\bibitem{yan2019meta}
Xiaopeng Yan, Ziliang Chen, Anni Xu, Xiaoxi Wang, Xiaodan Liang, and Liang Lin.
\newblock Meta r-cnn: Towards general solver for instance-level low-shot
  learning.
\newblock In {\em Proceedings of the IEEE/CVF International Conference on
  Computer Vision}, pages 9577--9586, 2019.

\bibitem{zhu2021semantic}
Chenchen Zhu, Fangyi Chen, Uzair Ahmed, and Marios Savvides.
\newblock Semantic relation reasoning for shot-stable few-shot object
  detection.
\newblock {\em arXiv preprint arXiv:2103.01903}, 2021.

\bibitem{SaligramaZero}
Pengkai Zhu, Hanxiao Wang, and Venkatesh Saligrama.
\newblock Zero shot detection.
\newblock {\em IEEE Transactions on Circuits and Systems for Video Technology},
  30(4):998--1010, 2019.

\end{thebibliography}
}

\appendix
\onecolumn
\maketitle
\section{Algorithm Introduction}
\begin{algorithm}  
  \caption{Metric of Category-based Grouping Mechanism}  
  \KwIn{The number of category for current Dataset is $N$; Category-based meta-features set $X$; All categories are grouped into $K$ groups by category-based grouping mechanism, and grouping results $C$=$\{C_1,...,C_K \}$; Constant $\tau$, $\epsilon$;}  
  \KwOut{$L_{re-meta}(X)$}  
   The mean value is $U=\{u_1,...,u_N\}$\; 
   The variance value is $\delta=\{\delta_1, ..., \delta_N\}$\; 
   $X\in \mathbb{R}^{N\times 1024}$ and $U,\delta \in \mathbb{R}^N$\;
   $C_i\cap C_j=\emptyset$ for $i\ne j, i,j\in \{1,2,...,K\}$\; 
   The constraint between groups is $L_{mean-std}$\;
   \For{$i=1; i \le N$}
   {
   	Compute the mean $u_i$ and variance $\delta_i$ of the meta-features using $X_i$\;
   }
   For all groups, the mean set is $W_{mean}$, the variance set is $W_{std}$, and the variance of mean is $W_{mean-std}$\;
  \For{$j=1;j \le K$}  
  {  
    Compute $u^j$ and $\delta^j$ of the $jth$ group using $C_j$ in Equation(8)\;
    \If{number of $C_j$ =1}
    {
    	Construct ${W_{mean}}^j$ by $C_j=\{m\}$ and $U$ in Equation(7)\;
    	Construct ${W_{std}}^j$ by $C_j=\{m\}$ and $\delta$ in Equation(6)\;
    }
    \Else
    {
      Compute ${W_{mean}}^j$ using $C_j$, $u^j$ and $U$ in Equation(7)\;
      Compute ${W_{std}}^j$ using $C_j$, $\delta^j$ and $\delta$ in Equation(6)\;
    }
     Compute ${W_{mean-std}}^j$ using $u^j$, $U$ and $C_j$ in Equation(5)\;
    }
    \For{$j=1;j\le K;$}
    {
       \For{$k=j+1; k\le K$}
       {
            ${L_{mean-std}}^j \leftarrow $ Compute $L_{j,k}$ by $W_{mean}$ and $W_{std}$ in Equation(4)\;
        }
        Compute $L^j_{group}$ using ${W_{mean-std}}^j$ and ${L_{mean-std}}^j$ in Equation(4)\;
       $L_{re-meta}(X) \leftarrow log(\tau+L^j_{group}$) in Equation(3)\; 
  }  
  return $L_{re-meta}(X)$\;  
\end{algorithm}

\section{Results and Visualization on Pascal VOC} 
 For novel set 1, 2, 3, this section shows our all results for $k$-shot detection, $k=1, 2, 3, 5.$ Finally, we visualize all results on all novel categories (i.e., boat, cat, mbike, sheep and sofa) for novel set1 2-shot.
\begin{table*}
 \small
 \centering
\resizebox{1.0\textwidth}{!}{
 \begin{tabular}{|ll|llllll|llllllllllllllll|}
    \hline
    \centering
   & &\multicolumn{6}{c|}{Novel} & \multicolumn{16}{c|}{Base}\\
     \hline
     \centering
    Shot&Method&boat&cat&mbike&sheep&sofa&mean&aero&bike&bird&bottle&bus&car&chair&cow&table&dog&horse&person&plant&train&tv&mean \\
    \hline
    \multirow{8}{*}{1}&YOLO-joint&0.0&9.1&0.0&0.0&0.0&1.8&78.7&76.8&73.4&48.8&79.0&82.3&50.2&68.4&71.4&76.7&80.7&75.0&46.8&83.8&71.7&70.9\\
    &YOLO-ft&0.1&25.8&10.7&3.6&0.1&8.1&77.2&74.9&69.1&47.4&78.7&79.7&47.9&68.3&69.6&74.7&79.4&74.2&42.2&82.7&71.1&69.1\\
    &YOLO-ft-full&0.1&30.9&26.0&8.0&0.1&13.0&75.1&70.7&65.9&43.6&78.4&79.5&47.8&68.7&68.0&72.8&79.5&72.3&40.1&80.5&68.6&67.4\\
    &\textbf{Baseline}&10.8&44.0&17.8&18.1&5.3&19.2&77.1&71.8&66.3&40.4&75.2&77.8&50.1&54.6&66.8&69.1&78.3&68.1&41.9&80.6&70.3&65.9\\
  &BCEwithLogits&9.09&22.3&25.47&15.56&9.67&16.42&71.91&73.19&62.09&42.19&72.81&76.81&46.55&51.06&63.8&66.6&77.99&67.24&40&80.42&69.46&64.11\\
 
  &Re-BCEwithLogits&2.6&23.81&19.89&19.76&0.22&13.26&71.61&72.1&66.8&41.64&74.47&75.84&46.48&57.61&66.5&68.8&79.13&70.27&40.78&78.62&69.68&65.36\\

    &Focal&1.55&\textbf{42.88}&15.23&16.72&4.98&16.27&75.89&72.62&71.47&43.01&78.7&76.87&48.79&53.16&62.29&71.2&79.33&71.7&38.22&77.16&68.65&65.94\\
   &Re-Focal&\textbf{9.79}&37.99&20.64&\textbf{20.14}&2.54&18.22&75.27&71.74&66.44&41.01&75.73&76.07&46.39&52.88&63.02&69.68&81.9&70.64&38.75&80.53&68.17&65.22\\

   &Cross-Entropy&3.52&29.01&25.6&16.43&2.27&15.37&71.6&72.08&65.53&39.42&75.42&76.06&43.93&57.07&64.87&68.09&78.69&69.24&37.81&78.67&68.35&64.45\\
  &Re-Cross-Entropy&9.52&33.67&27.33&12.35&9.85&18.55&73.86&74.4&66.14&40.42&73.87&76.45&47.09&53.91&68.13&70.78&78.2&67.53&35.64&76.24&67.65&64.69\\

  &TCL-2 &6.43&33.8&31.21&11.51&\textbf{12.8}&19.15&72.03&71&67.24&42.16&74.23&76.77&46.57&54.34&65.69&71.61&81.02&69.77&41.35&77.03&69.46&65.35\\
  &\textbf{Ours}&9.53&33.58&\textbf{32.28}&19.66&5.34&\textbf{20.08}&71.98&72.65&65.45&41.96&75.14&78.44&43.75&52.03&65.04&73.27&80.29&68.78&43.46&80.36&68.97&65.44\\
    \hline
        
  \multirow{8}{*}{2}&YOLO-joint &0.0&9.1&0.0&0.0&0.0&1.8&77.6&77.1&74.0&49.4&79.8&79.9&50.5&71.0&72.7&76.3&81.0&75.0&48.4&84.9&72.7&71.4\\
  &YOLO-ft&0.0&24.4&2.5&9.8&0.1&7.4&78.2&76.0&72.2&47.2&79.3&79.8&47.3&72.1&70.0&74.9&80.3&74.3&45.2&84.9&72.0&70.2\\
  &YOLO-ft-full&0.0&35.2&28.7&15.4&0.1&15.9&75.3&72.0&69.8&44.0&79.1&78.8&42.1&70.0&64.9&73.8&81.7&71.4&40.9&80.9&69.4&67.6\\
  &\textbf{Baseline}&5.3&46.4&18.4&26.1&12.4&21.7&71.4&72.4&64.5&37.9&75.3&77.1&42.9&55.0&57.4&73.7&78.9&68.0&41.5&75.9&69.0&64.1\\
   &BCEwithLogits&4.57&30.21&23.74	&17.66&16.36&18.51&69.8&71.62&62.79&40.77&73.2&77.57&	46.19&55.03&64.48&67.55&78.61&68.79&39.37&74.09&66.16	&63.74\\
  &Re-BCEwithLogits&2.93&28.84&19.64&27.16&8.74&17.46&71.5&72.77&68.1&40.92&74.89&75.99&42.6&58.7&65.78&67.99&79.35&69.94&40.27&73.78&66.97&64.64\\

    &Focal&2.58&\textbf{43.85}&11.29&33.16&17.29&21.63&72.16&71.4&69.1&42.31&75.8&73.5&44.85&61.52&64.11&70.09&79.28&70&39.02&78.23&69.14&65.37\\
   &Re-Focal&6.15&33.6&18.95&29.43&12.1&20.05&74.88&62.18&66.6&41.13&76.91&76.51&45.4&60.88&62.23&70.71&82.32&71.68&42.19&74.26&67.51&65.03\\

   &Cross-Entropy&10.25&36.46&16.2&24.97&7.66&19.11&72.31&75.59&67.87&40.79&74.26&76.95&42.75&59.03&66.24&72.86&79.03&71.4&39.73&75.37&67.32&65.43\\
  &Re-Cross-Entropy&10.06&38.83&19.41&17.65&\textbf{19.15}&21.02&70.92&73.33&66.49&38.73&74.36&75.59&46.96&59.51&66.33&70.34&78.13&70.29&40.36&69.74&67.04&64.54\\

  &TCL-2 &4.95&32.49&28.31&27.06&13.34&21.23&60.79&70.66&67.51&39.37&74.61&76.4&39.26&54.21&66.68&70.81&79.91&69.22&41.68&67.71&67.89&63.71\\
  &\textbf{Ours}&\textbf{10.61}&35.11&\textbf{33.75}&\textbf{35.89}&18.38&\textbf{26.75}&70.53&73.53&66.6&43.18&75.52&77.71&41.29&53.9&63.63&72.5&80.85&69.77&43.08&77.27&68.42&65.18\\
   \hline  
    
  \multirow{8}{*}{3}&YOLO-joint&0.0&9.1&0.0&0.0&0.0&1.8&77.1&77.0&70.6&46.3&77.5&79.7&49.7&68.8&73.4&74.5&79.4&75.6&48.1&83.6&72.1&70.2\\
  &YOLO-ft&0.0&27.0&1.8&9.1&0.1&7.6&77.7&76.6&71.4&47.5&78.0&79.9&47.6&70.0&70.5&74.4&80.0&73.7&44.1&83.0&70.9&69.7\\
  &YOLO-ft-full&0.0&39.0&18.1&17.9&0.0&15.0&73.2&71.1&68.8&43.7&78.9&79.3&43.1&67.8&62.2&76.3&79.4&70.8&40.5&81.6&69.6&67.1\\
  &\textbf{Baaseline}&11.2&39.8&20.9&23.7&33.0&25.7&73.2&68.0&65.9&39.8&77.3&77.5&43.5&57.7&60.7&64.5&77.5&68.4&42.0&80.6&70.2&64.4\\
  &BCEwithLogits&\textbf{10.92}&\textbf{46.62}&24.57&23.23&31.71&27.41&73.06&71.22&62.71&40.1&73.9&77.79&46.02&55.06&62.75&67.95&79.51&65.49&39.81&76.69&67.36&63.96\\
  &Re-BCEwithLogits&9.77&45.18&22.77&22.37&21.47&24.31&71.95&70.14&63.95&38.09&74.65&74.86&41.02&57.12&58.25&63.08&79.29&65.36&40.36&73.38&67.43&62.6\\

    &Focal&10.36&43.77&19.45&28.41&\textbf{37.56}&27.91&72.71&71.66&66.84&42.79&76.13&76.71&44.92&57.42&66.61&68&81.97&68.8&40.25&77.19&67.44&65.3\\
   &Re-Focal&11.67&27.17&16.83&30.07&16.52&20.45&73.75&72.52&69.66&43.83&76.44&77.94&48.14&64.6&66.78&71.62&83.64&70.66&43.86&79.86&68.55&67.32\\

   &Cross-Entropy&7.42&32.57&14.75&\textbf{31.04}&29.76&23.11&70.8&73.39&69.21&40.52&75.33&78.98&46.07&57.42&66.21&69.44&79.53&69.03&38.8&79.15&65.43&65.29\\
  &Re-Cross-Entropy&10.47&32.89&16.06&24.1&28.75&22.25&74.1&74.44&68&41.23&77.62&76.84&48.19&58.56&67.15&69.26&78.71&69.84&38.12&80.31&67.1&65.96\\

  &TCL-2 &9.67&47.2&25.11&28.97&32.26&28.64&72.5&67.82&67.63&39.08&75.28&77.06&43.79&53.3&65.71&72.9&80.65&69.11&41.24&76.85&69.16&64.8\\
  &\textbf{Ours}&10.29&46.05&\textbf{28.11}&29.81&34.52&\textbf{29.76}&72.45&73.14&66.32&41.18&75.07&78.55&45.04&56.81&64.46&72.73&81.23&68.26&42.61&79.87&66.89&65.64\\
   \hline
        
  \multirow{8}{*}{5}&YOLO-joint&0.0&9.1&0.0&0.0&9.1&3.6&78.2&78.5&72.1&47.8&76.6&82.1&50.7&70.1&71.8&77.6&80.4&75.4&46.0&84.8&72.5&71.0\\
  &YOLO-ft&0.0&33.8&2.6&7.8&3.2&9.5&77.2&77.1&71.9&47.3&78.8&79.8&47.1&69.8&71.8&77.0&80.2&74.3&44.2&82.5&70.6&70.0\\
  &YOLO-ft-full&7.9&48.0&39.1&29.4&36.6&32.2&75.5&73.6&69.1&43.3&78.4&78.9&42.3&70.2&66.1&77.4&79.8&72.2&41.9&82.8&69.3&68.1\\
  &\textbf{Baseline}&\textbf{14.2}&\textbf{57.3}&\textbf{50.8}&38.9&41.6&\textbf{40.6}&70.1&66.3&66.5&40.0&78.1&77.0&40.4&61.2&61.5&71.2&79.1&70.4&38.5&80.0&68.0&64.6\\
  &BCEwithLogits&8.91&49.65&49.11&31.72&40.95&36.07&70.68&73.05&65.16&38.42&75&77.82&41.25&61.02&62.72&71.88&79.94&68.33&40.99&77.75&67.68&64.78\\
  &Re-BCEwithLogits&12.61&44.16&42.49&36.93&32.59&33.76&70.83&73.53&67.28&38.7&76.79&76.65&39.31&64.76&63.29&69.56&81.21&66.78&39.42&78.28&68.1&64.97\\

    &Focal&8.04&52.42&48.07&39.9&38.76&37.43&72.2&72.54&66.31&40.65&78.69&76.74&43.54&64.5&68.53&69.94&81.07&69.84&37.68&80.03&69.13&66.09\\
   &Re-Focal&12.66&42.89&43.28&\textbf{40.7}&\textbf{43.44}&36.15&73.2&66.09&66.03&41.02&77.06&76.78&44.6&65.74&63.48&67.93&82.62&70.46&40.7&76.88&68.1&65.38\\

   &Cross-Entropy&7.37&45.65&44.55&39.62&38.7&35.18&72.32&69.03&67.67&40.15&77.41&77.16&40.29&60.94&64.86&73.51&81.17&69.68&38.94&80.75&67.45&65.42\\
  &Re-Cross-Entropy&10.66&48.92&44.25&36.05&42.65&36.5&75.5&75.17&69.9&40.72&77.28&76.96&47.54&62.26&65.63&73.91&79.66&70.86&38.87&80.68&68.61&66.9\\

  &TCL-2 &5.94&55.34&49.25&34.84&39.33&36.94&70.44&66.41&64.71&35.75&75.56&75.54&35.72&59.05&57.86&73.47&77.87&66.64&37&75.44&65.73&62.48\\
  &\textbf{Ours}&9.02&47.13&\textbf{49.78}&40.11&35.37&36.28&73.28&74.77&67.25&39.4&76.51&78.53&44.52&57.32&66.44&75.54&81.33&58.92&39.8&80.15&68.98&66.18\\    
  \hline 
 \end{tabular}
 }  
\label{novel_set1}
\caption{The results of detection APs (\%). For few-shot detection on Pascal VOC, ours significantly outperforms others on novel set1.}
\end{table*}
\begin{table*}
 \small
 \centering
  
\resizebox{1.0\textwidth}{!}{
 \begin{tabular}{|ll|llllll|llllllllllllllll|}
    \hline
    \centering
   & &\multicolumn{6}{c|}{Novel} & \multicolumn{16}{c|}{Base}\\
     \hline
     \centering
    Shot&Method&bird&bus&cow&mbike&sofa&mean&aero&bike&boat&bottle&car&cat&chair&table&dog&horse&person&plant&sheep&train&tv&mean\\
    \hline
  \multirow{8}{*}{1}&YOLO-joint&0.0&0.0&0.0&0.0&0.0&0.0&78.4&76.9&61.5&48.7&79.8&84.5&51.0&72.7&79.0&77.6&74.9&48.2&62.8&84.8&73.1&70.2\\
  &YOLO-ft&6.8&0.0&9.1&0.0&0.0&3.2&77.1&78.2&61.7&46.7&79.4&82.7&51.0&69.0&78.3&79.5&74.2&42.7&68.3&84.1&72.9&69.7\\
  &YOLO-ft-full&11.4&17.6&3.8&0.0&0.0&6.6&75.8&77.3&63.1&45.9&78.7&84.1&52.3&66.5&79.3&77.2&73.7&44.0&66.0&84.2&72.2&69.4\\
  &\textbf{Baseline}&13.5&10.6&31.5&13.8&4.3&14.8&75.1&70.7&57.0&41.6&76.6&81.7&46.6&72.4&73.8&76.9&68.8&43.1&63.0&78.8&69.9&66.4\\
  &BCEwithLogits&10.56&17.23&5.64&\textbf{33.62}&0.91&13.59&74.13&69.35&54.19&38.89&75.62&80.05&48.07&61.85&72.03&74.75&64.59&38.88&58.63&77.72&68.27&63.8\\
  &Re-BCEwithLogits&5.89&38.39&14.65&32.00&0.51&\textbf{18.29}&73.1&69.84&55.76&40.42&76.81&78.51&46.1&63.73&75.78&77.92&68.69&40.41&59.85&79.28&70.79&65.13\\

    &Focal&11.18&1.17&13.66&25.83&0.1&10.39&74.61&73.96&53.91&41.82&53.8&78.28&48.87&24.36&78.16&76.08&71.31&42.12&64.13&81.33&70.97&62.25\\
   &Re-Focal&11.41&6.89&22.09&28.55&1.89&14.16&72.78&69.65&55.88&40.02&77.14&82.69&47.32&67.15&78.6&77.33&68.6&40.36&62.98&80.14&70.69&66.09\\

   &Cross-Entropy&12.49&11.16&\textbf{24.72}&28.79&3.03&16.04&72.85&69.98&55.87&43.31&76.53&77.72&45.37&69.87&75.76&75.98&69.72&40.07&59.27&78.15&70.13&65.37\\
  &Re-Cross-Entropy&12.35&16.28&13.23&33.41&0.47&15.15&73.02&70.19&60.02&41.78&77.59&80.05&44.86&66.17&78.08&75.95&70&40&59.47&79.68&68.27&65.68\\

  &TCL-2 &18.21&18.02&15.38&24.2&\textbf{11.99}&17.56&69.95&72.15&55.51&37.15&76.74&81.57&45.66&68.99&72.95&75.79&68.26&41.47&62.86&78.64&70.91&65.24\\
  &\textbf{Ours}&\textbf{19.18}&\textbf{22.58}&7.62&30.81&10.16&18.07&72.43&76.21&56.25&40.45&77.07&81.02&47.89&67.63&75.06&74.2&69.72&39.38&61.51&76.96&67.46&65.55\\
   \hline
        
  \multirow{8}{*}{2} &YOLO-joint&0.0&0.0&0.0&0.0&0.0&0.0&77.6&77.6&60.4&48.1&81.5&82.6&51.5&72.0&79.2&78.8&75.2&47.0&65.2&86.0&72.7&70.4\\
  &YOLO-ft&11.5&5.8&7.6&0.1&7.5&6.5&77.9&75.0&58.5&45.7&77.6&84.0&50.4&68.5&79.2&79.7&73.8&44.0&66.0&77.5&72.9&68.7\\
  &YOLO-ft-full&16.6&9.7&12.4&0.1&14.5&10.7&76.4&70.2&56.9&43.3&77.5&83.8&47.8&70.7&79.1&77.6&71.7&39.6&61.4&77.0&70.3&66.9\\
  &\textbf{Baseline}&21.2&12.0&16.8&17.9&9.6&15.5&74.6&74.9&56.3&38.5&75.5&68.0&43.2&69.3&66.2&42.4&68.1&41.8&59.4&76.4&70.3&61.7\\
   &BCEwithLogits&19.22&12.11&21.55&14.98&5.7&14.71&71.83&70.38&54.28&37.96&75.46&81.24&44.12&71.37&77.53&76.8&69.95&39.44&54.8&72.74&66.68&64.3\\
  &Re-BCEwithLogits&15.18&6.49&27.69&26.89&\textbf{22.27}&19.71&72.21&72.97&55.33&39.56&74.71&77.94&43.68&65.32&77.86&77.43&68.3&39.71&57.31&76.16&68.11&64.44\\

    &Focal&19.13&11.25&28.13&15.09&2.56&15.23&74.46&77.02&54.96&43.77&77.86&82.33&45.03&68.47&79.01&75.34&72.02&43.38&62.98&79.66&68.55&66.99\\
   &Re-Focal&12.45&3.88&37.62&19.14&6.28&15.88&72.29&72.52&59.1&44.79&78.15&82.47&47.14&68.25&79.49&76.44&72.27&41.79&67.79&75.44&69.9&67.19\\

   &Cross-Entropy&24.64&7.22&27.93&27.00&9.19&19.2&74.52&70.24&55.12&42.94&77.14&78.71&44.82&69.75&76.92&69.73&69.15&41.45&58.8&73.04&68.76&64.74\\
  &Re-Cross-Entropy&23.43&11.61&19.13&\textbf{37.85}&12.04&20.81&71.79&70.63&59.38&41.65&77.42&81.35&44.56&66.39&78.41&56.23&70.88&40.77&52.71&73.85&68.52&63.64'\\

  &TCL-2 &23.02&8.3&\textbf{36.51}&27.34&16.09&22.25&72.65&70.28&54.02&36.6&76.41&80.82&42.55&67.26&75.97&77.13&69.15&41.8&60.92&73.06&68.71&64.49\\
  &\textbf{Ours}&\textbf{25.18}&\textbf{12.29}&31.87&32.71&21.23&\textbf{24.66}&72.31&73.62&55.07&38.23&74.73&81.85&44.37&65.41&75.94&75.85&69.93&39.81&56.93&73.31&65.72&64.21\\
    \hline
    
  \multirow{8}{*}{3} &YOLO-joint&0&0&0&0&9.1&1.8&78.0&77.2&61.2&45.6&81.6&83.7&51.7&73.4&80.7&79.6&75.0&45.5&65.6&83.1&72.7&70.3\\
  &YOLO-ft&10.9&5.5&15.3&0.2&0.1&6.4&76.7&77.0&60.4&46.9&78.8&84.9&51.0&68.3&79.6&78.7&73.1&44.5&67.6&83.6&72.4&69.6\\
  &YOLO-ft-full&21.0&22.0&19.1&0.5&0.0&12.5&73.4&67.5&56.8&41.2&77.1&81.6&45.5&62.1&74.6&78.9&67.9&37.8&54.1&76.4&71.9&64.4\\
  &\textbf{Baseline}&26.1&19.1&40.7&20.4&27.1&26.7&73.6&73.1&56.7&41.6&76.1&78.7&42.6&66.8&72.0&77.7&68.5&42.0&57.1&74.7&70.7&64.8\\
  &BCEwithLogits&24.71&22.63&29.19&\textbf{29.22}&25.75&26.3&71.62&70.05&54.66&37.51&75.44&79.7&46.09&63.79&75.58&77.37&67.16&40.1&49.63&73.3&66.12&63.21\\
  &Re-BCEwithLogits&23.03&\textbf{29.91}&22.74&19.06&40.2&26.99&70.37&72.92&56.98&37.24&75.72&79.38&45.12&66.07&74.99&78.59&65.85&40.35&49.45&77.29&68.68&63.93\\

    &Focal&10.3&16.38&33.76&17.11&14.24&18.36&73.33&74.27&55.78&43.21&79.32&82.48&47.69&70.4&76.97&79.0&70.32&42.39&61.53&81.29&70.48&67.23\\
   &Re-Focal&25.98&11.44&37.02&20.43&20.79&23.13&72.58&70.06&58.43&42.27&78.28&80.97&47.52&66.97&78.47&76.03&68.9&41.7&60.5&75.51&69.02&65.81\\

   &Cross-Entropy&28.17&14.07&32.98&24.2&27.86&25.46&73.07&70.39&56.28&41.02&73.88&80.17&45.97&70.5&75.79&77.58&68.22&40.2&55.15&76.83&69.07&64.94\\
  &Re-Cross-Entropy&23.25&18.64&\textbf{37.54}&24.33&26.62&26.07&70.03&71.94&56.92&40.92&77.53&81.16&47.0&68.24&76.68&77.33&69.29&41.12&51.47&77.53&69.22&65.09\\

  &TCL-2 &31.15&23.29&23.97&19.96&29.49&25.57&67.69&66.26&51.65&35.66&76.84&80.62&45.79&66.01&74.73&77.15&65.55&40.55&57.96&77.71&69.72&63.59\\
  &\textbf{Ours}&\textbf{32.1}&20.53&30.64&28.52&\textbf{42.9}&\textbf{30.94}&71.19&69.39&51.81&37.66&76.41&81.63&45.39&64.51&75.41&76.11&69.45&39.09&54.88&75.38&67.9&63.75\\
   \hline
        
  \multirow{8}{*}{5}&YOLO-joint&0.0&0.0&0.0&0.0&9.1&1.8&77.8&76.4&65.7&45.9&79.5&82.3&50.4&72.5&79.1&79.0&75.5&47.9&67.2&83.0&72.5&70.3\\
&YOLO-ft&11.6&7.1&10.7&2.1&6.0&7.5&76.5&76.4&61.0&45.5&78.7&84.5&49.2&68.7&78.5&78.1&73.7&45.4&66.8&85.3&70.0&69.2\\
&YOLO-ft-full&20.2&20.0&22.4&36.4&24.8&24.8&72.0&70.6&60.7&42.0&76.8&84.2&47.7&63.7&76.9&78.8&72.1&42.2&61.1&80.8&69.9&66.6\\
&\textbf{Baseline}&31.5&21.1&39.8&40.0&37.0&33.9&69.3&57.5&56.8&37.8&74.8&82.8&41.2&67.3&74.0&77.4&70.9&40.9&57.3&73.5&69.3&63.4\\

  &BCEwithLogits&30.15&24.25&32.32&52.43&36.83&35.2&70.62&71.46&54.45&38.34&75.34&80.7&47.78&65.24&73.84&77.03&65.86&36.92&48.75&73.89&65.78&63.07\\
  &Re-BCEwithLogits&25.41&36.34&25.49&46.96&\textbf{42.29}&35.3&69.14&74.33&55.5&37.18&76.77&79.12&45.9&64.79&76.47&77.61&67.74&38.64&53.98&78.21&68.97&64.29\\

    &Focal&28.07&20.89&40.79&49.65&31.03&34.09&73.52&68.13&56.93&39.23&76.85&80.54&44.15&67.99&77.37&75.21&69.25&40.05&59.0&78.25&70.47&65.13\\
   &Re-Focal&18.82&21.87&38.92&33.81&22.6&27.2&73.12&73.52&60.79&42.24&78.96&83.06&49.59&68.32&79.09&77.7&71.54&41.5&66.99&80.1&70.96&67.83\\

   &Cross-Entropy&\textbf{33.26}&25.01&\textbf{40.64}&38.19&42.11&35.84&71.91&70.27&58.51&40.01&76.6&80.33&45.54&69.93&78.19&77.44&70.96&40.05&58.74&80.47&69.78&65.92\\
  &Re-Cross-Entropy&29.4&14.81&37.89&49.28&35.85&33.45&70.98&70.31&56.94&41.35&77.53&83.88&45.11&67.13&76.83&78.14&70.17&41.45&56.7&76.74&67.41&65.38\\

  &TCL-2 &29.84&\textbf{42.46}&30.11&48.26&41.58&38.45&67.27&60.54&52.2&32.59&74.41&81.03&33.61&62.9&67.29&75.62&64.23&32.44&56.2&71.65&66.99&59.93\\
  &\textbf{Ours}&30.35&37.51&30.7&\textbf{55.16}&41.49&\textbf{39.04}&67.82&67.17&48.47&33.66&73.42&78.18&39.45&61.11&69.54&74.0&68.07&36.5&55.67&71.31&65.04&60.63\\  
    \hline 
 \end{tabular}
 }
\label{novel_set2}
\caption{The results of detection APs (\%). For few-shot detection on Pascal VOC, ours significantly outperforms others on novel set2.}
\end{table*}
\begin{table*}
 \small
 \centering
 
\resizebox{1.0\textwidth}{!}{
 \begin{tabular}{|ll|llllll|llllllllllllllll|}
    \hline
    \centering
   & &\multicolumn{6}{c|}{Novel} & \multicolumn{16}{c|}{Base}\\
    \hline
     \centering
    Shot&Method&aero&bottle&cow&horse&sofa&mean&bike&bird&boat&bus&car&cat&chair&table&dog&mbike&person&plant&sheep&train&tv&mean\\
    \hline
  \multirow{8}{*}{1}&YOLO-joint&0.0&0.0&0.0&0.0&0.0&0.0&78.8&73.2&63.6&79.0&79.7&87.2&51.5&71.2&81.1&78.1&75.4&47.7&65.9&84.0&73.7&72.7\\
  &YOLO-ft&0.4&0.2&10.3&29.8&0.0&8.2&77.9&70.2&62.2&79.8&79.4&86.6&51.9&72.3&77.1&78.1&73.9&44.1&66.6&83.4&74.0&71.8\\
  &YOLO-ft-full&0.6&9.1&11.2&41.6&0.0&12.5&74.9&67.2&60.1&78.8&79.0&83.8&50.6&72.7&75.5&74.8&71.7&43.9&62.5&81.8&72.6&70.0\\
  &\textbf{Baseline}&11.8&9.1&15.6&23.7&18.2&15.7&77.6&62.7&54.2&75.3&79.0&80.0&49.6&70.3&78.3&78.2&68.5&42.2&58.2&78.5&70.4&68.2\\
  &BCEwithLogits&10.12&0.7&7.86&39.72&\textbf{17.1}&15.1&74.81&64.83&52.22&73.61&76.19&78.31&46.72&66.34&78.4&77.34&67.81&37.57&58.53&74.82&70.73&66.55\\
  &Re-BCEwithLogits&10.74&0.51&7.39&31.33&5.05&11&73.86&62.79&49.72&72.46&77.28&80.37&44.56&66.54&76.61&75.87&65.98&38.26&57.72&71.58&68.98&65.5\\

    &Focal&10.76&0.06&16.13&19.98&1.06&9.6&75.75&65.05&57.9&77.11&77.29&81.48&46.04&62.73&75.47&74.19&69.38&40.55&54.75&78.08&69.18&67\\
   &Re-Focal&9.48&0.03&19.36&7.48&0.13&7.3&78.53&68.12&59.02&77.91&78.58&83.09&50.56&57.57&72.67&76.19&69.51&42.75&59.86&72.39&70.87&67.8\\

   &Cross-Entropy&14.01&0.34&10.89&25.89&9.88&12.19&73.46&62.63&52.58&74.92&76.8&81.12&46.33&66.23&79.3&74.31&68.15&38.79&56.06&76.86&69.38&66.46\\
  &Re-Cross-Entropy&9.28&0.22&\textbf{21.57}&25.2&9.09&13.07&75.33&64.84&54.94&75.81&77.84&82.48&47.64&63.8&78.3&75.53&70.4&38.38&57.96&76.01&67.7&67.13\\

  &TCL-2 &0.76&0.07&16.96&29.59&13.99&12.27&75.82&60.6&54.09&75.65&77.95&84.08&48.81&64.8&73.36&77.1&71.26&40.53&58.01&78.17&69.27&67.3\\
  &\textbf{Ours}&\textbf{16.33}&\textbf{9.09}&8.7&\textbf{46.9}&16.08&\textbf{19.42}&72.71&62.94&45&74.97&75.29&77.39&44.73&64.71&67.84&74.99&68.65&38.36&54.3&76.09&68.95&64.46\\
    \hline
        
  \multirow{8}{*}{2}&YOLO-joint&0.0&0.6&0.0&0.0&0.0&0.1&78.4&69.7&64.5&78.3&79.7&86.1&52.2&72.6&81.2&78.6&75.2&50.3&66.1&85.3&74.0&72.8\\
  &YOLO-ft&0.2&0.2&17.2&1.2&0.0&3.8&78.1&70.0&60.6&79.8&79.4&87.1&49.7&70.3&80.4&78.8&73.7&44.2&62.2&82.4&74.9&71.4\\
  &YOLO-ft-full&1.8&1.8&15.5&1.9&0.0&4.2&76.4&69.7&58.0&80.0&79.0&86.9&44.8&68.2&75.2&77.4&72.2&40.3&59.1&81.6&73.4&69.5\\
  &\textbf{Baseline}&28.6&0.9&27.6&0.0&19.5&15.3&75.8&67.4&52.4&74.8&76.6&82.5&44.5&66.0&79.4&76.2&68.2&42.3&53.8&76.6&71.0&67.2\\
    &BCEwithLogits&28.39&0.61&28.6&0.53&\textbf{19.97}&15.62&74.24&67.99&51.14&75.32&76.18&79.04&45.98&66.83&77.66&77.48&68.75&38.91&55.13&75.69&70.05&66.69\\
  &Re-BCEwithLogits&21.59&0.22&25.02&0.19&17.96&13&75.95&64.26&48.5&76.34&77.12&81.84&43.76&67.99&78.27&76.22&68.65&42.37&54.22&76.54&68.03&66.67\\

    &Focal&26.74&\textbf{1.3}&13.04&0.66&2.59&8.87&73.85&67.33&54.91&77.18&78.56&82.54&44.52&63.84&80.36&77.87&69.33&42.3&58.35&77.97&69.95&67.92\\
   &Re-Focal&14.44&0.29&24.55&0.13&3.94&8.67&77.32&70.66&56.19&78.09&78.07&83.41&48.05&65.54&79.71&78.13&68.42&42.57&59.41&76.83&70.55&68.86\\

   &Cross-Entropy&28.6&0.9&27.6&0&19.5&15.3&75.8&67.4&52.4&74.8&76.6&82.5&44.5&66&79.4&76.2&68.2&42.3&53.8&76.6&71&67.2\\
  &Re-Cross-Entropy&28.4&0.13&28.42&0.57&15.15&14.53&75.4&67.05&57.57&75.18&77.82&82.17&49.28&71.61&79.59&76.72&70.17&42.48&59.89&76.29&68.35&68.64'\\

  &TCL-2 &\textbf{34.66}&0.83&31.32&1.01&18.82&17.33&69.88&61.47&50.09&72.15&73.96&77.26&38.45&61.21&73.48&74.39&66.28&37&50.94&71.34&66.07&62.93\\
  &\textbf{Ours}&34.14&0.49&\textbf{34.6}&\textbf{2.27}&15.65&\textbf{17.43}&77.56&65.59&51.74&72.95&76.14&79.49&42.88&64.67&76.65&75.98&70.37&41.08&55.8&75.15&68.4&66.3\\
    \hline
    
  \multirow{8}{*}{3}&YOLO-joint&0.0&0.0&0.0&0.0&0.0&0.0&77.6&72.2&61.2&77.9&79.8&85.8&49.9&73.2&80.0&77.9&75.3&50.8&64.3&84.2&72.6&72.2\\
  &YOLO-ft&4.9&0.0&11.2&1.2&0.0&3.5&78.7&71.6&62.4&77.4&80.4&87.5&49.5&70.8&79.7&79.5&72.6&44.3&60.0&83.0&75.2&71.5\\
  &YOLO-ft-full&10.7&4.6&12.9&29.7&0.0&11.6&74.9&69.2&60.4&79.4&79.1&87.3&43.4&69.7&75.8&75.2&70.5&39.4&52.9&80.8&73.4&68.8\\
  &\textbf{Baseline}&29.4&4.6&34.9&6.8&37.9&22.7&62.6&64.7&55.2&76.6&77.1&82.7&46.7&65.4&75.4&78.3&69.2&42.8&45.2&77.9&69.6&66.0\\
  &BCEwithLogits&32.19&9.09&29.2&18.99&\textbf{41.25}&26.14&71.29&64.92&50.96&75.89&76.87&80.89&47.81&67.03&75.62&75.87&65.67&37.94&40.8&78.44&70.66&65.38\\
  &Re-BCEwithLogits&28.57&1.59&30.45&12.91&30.16&20.74&69.18&62.13&50.28&75.55&77.51&81.86&44.31&67.74&76.39&76.9&65.58&40.88&48.33&77.22&66.98&65.37\\

    &Focal&41.55&0.14&30.81&5.24&23.06&20.16&68.15&65.43&53.92&78.67&77.53&81.51&46.51&62.56&79.72&76.13&70.46&41.11&53.42&76.39&68.93&66.7\\
   &Re-Focal&23.68&0.14&30.79&11.23&15.91&16.35&71.85&69.68&52.25&78.67&78.78&83.68&50.77&66.81&78.45&76.73&70.02&40.98&56.54&76.72&71.39&68.22\\

   &Cross-Entropy&36.47&0.83&\textbf{35.42}&4.65&24.2&20.31&73.66&64.98&51.99&75.61&77.96&84.05&46.85&66.4&75.29&79.23&70.37&41.69&52.23&78.26&69.17&67.18\\
  &Re-Cross-Entropy&35.59&1.3&38.54&12.54&31.71&23.93&70.71&67.85&48.91&76.36&78.06&82.38&49.33&68.88&78.77&78.7&69.41&40.87&56.92&77.58&68.21&67.53\\

  &TCL-2 &\textbf{44.05}&\textbf{9.09}&29.74&\textbf{30.28}&40.9&\textbf{30.81}&64.75&60.22&50.98&73.25&74.53&75.68&42.18&59.55&69.3&73.94&62.4&37.19&42.58&71.69&67.77&61.73\\
  &\textbf{Ours}&43.57&3.03&26.28&8.77&34.58&23.24&74.94&65.33&53.83&73.41&77.16&80.42&46.47&62.36&74.49&77.22&68.61&42.38&56.93&78.05&69.73&66.76\\
   \hline
        
  \multirow{8}{*}{5}&YOLO-joint&0.0&0.0&0.0&0.0&9.1&1.8&78.0&71.5&62.9&81.7&79.7&86.8&50.0&72.3&81.7&77.9&75.6&48.4&65.4&83.2&73.6&72.6\\
  &YOLO-ft&0.8&0.2&11.3&5.2&0.0&3.5&78.6&72.4&61.5&79.4&81.0&87.8&48.6&72.1&81.0&79.6&73.6&44.9&61.4&83.9&74.7&72.0\\
  &YOLO-ft-full&10.3&9.1&17.4&43.5&0.0&16.0&76.4&69.6&59.1&80.3&78.5&87.8&42.1&72.1&76.6&77.1&70.7&43.1&58.0&82.4&72.6&69.8\\
  &\textbf{Baseline}&33.1&9.4&38.4&25.4&44.0&30.1&73.2&65.6&52.9&75.9&77.5&80.0&43.7&65.0&73.8&78.4&68.9&39.2&56.4&78.0&70.8&66.6\\
  &BCEwithLogits&34.95&9.6&39.77&38.42&35.25&31.6&62.67&60.27&47.94&73.84&76.02&81.24&36.22&58.34&70.32&76.77&64.08&34.21&44.07&67.14&68.85&61.46\\
  &Re-BCEwithLogits&39.54&\textbf{10.18}&29.8&39.16&41.09&31.95&68.47&62.8&50.71&75.66&75.9&82.76&38.52&64.22&66.51&76.59&65.42&36.26&47.94&77.55&66.9&63.75\\

    &Focal&36.72&9.25&38.45&19.4&33.85&27.54&71.35&63.31&50.51&78.66&77.23&82.49&43.95&60.1&76.13&77.55&70&37.36&53.92&77.12&69.18&65.92\\
   &Re-Focal&39.04&9.27&37.96&28.32&28.4&28.6&68.26&66.08&51.29&78.9&77.25&83.35&44.7&65.53&76.78&74.94&68.89&38.34&54.3&77.37&70.51&66.43\\

   &Cross-Entropy&37.34&9.86&34.27&38.2&39.86&31.91&70.22&59.53&47.57&74.56&75.95&82.2&43.24&65.92&68.43&76.71&68.36&35.22&50.16&77.82&67.55&64.23\\
  &Re-Cross-Entropy&38.61&9.65&\textbf{42.49}&39.82&\textbf{47.31}&35.58&67.01&62.41&52.4&74.6&74.65&80.34&42.02&67.56&73.28&75.78&67&38.57&50.48&74.02&67.88&64.53\\

  &TCL-2 &\textbf{47.64}&9.41&36.71&41.67&39.97&35.08&64.22&60.49&52.15&75.61&73.65&78.92&36.39&58.68&70.4&75.1&62.92&32.91&45.41&75.02&67.69&61.97\\
  &\textbf{Ours}&44.66&9.88&37.51&\textbf{55.49}&40.75&\textbf{37.66}&64.54&63.48&50.68&75.2&74.96&77.02&38.67&59.23&71.05&75.06&64.4&35.82&52.06&74.45&68.15&62.98\\  
    \hline 
 \end{tabular}
 }
\label{novel_set3}
\caption{The results of detection APs (\%). For few-shot detection on Pascal VOC, ours significantly outperforms others on novel set3.}
\end{table*}
\clearpage
\begin{figure*}
  \centering
  \includegraphics[width=1.0\columnwidth]{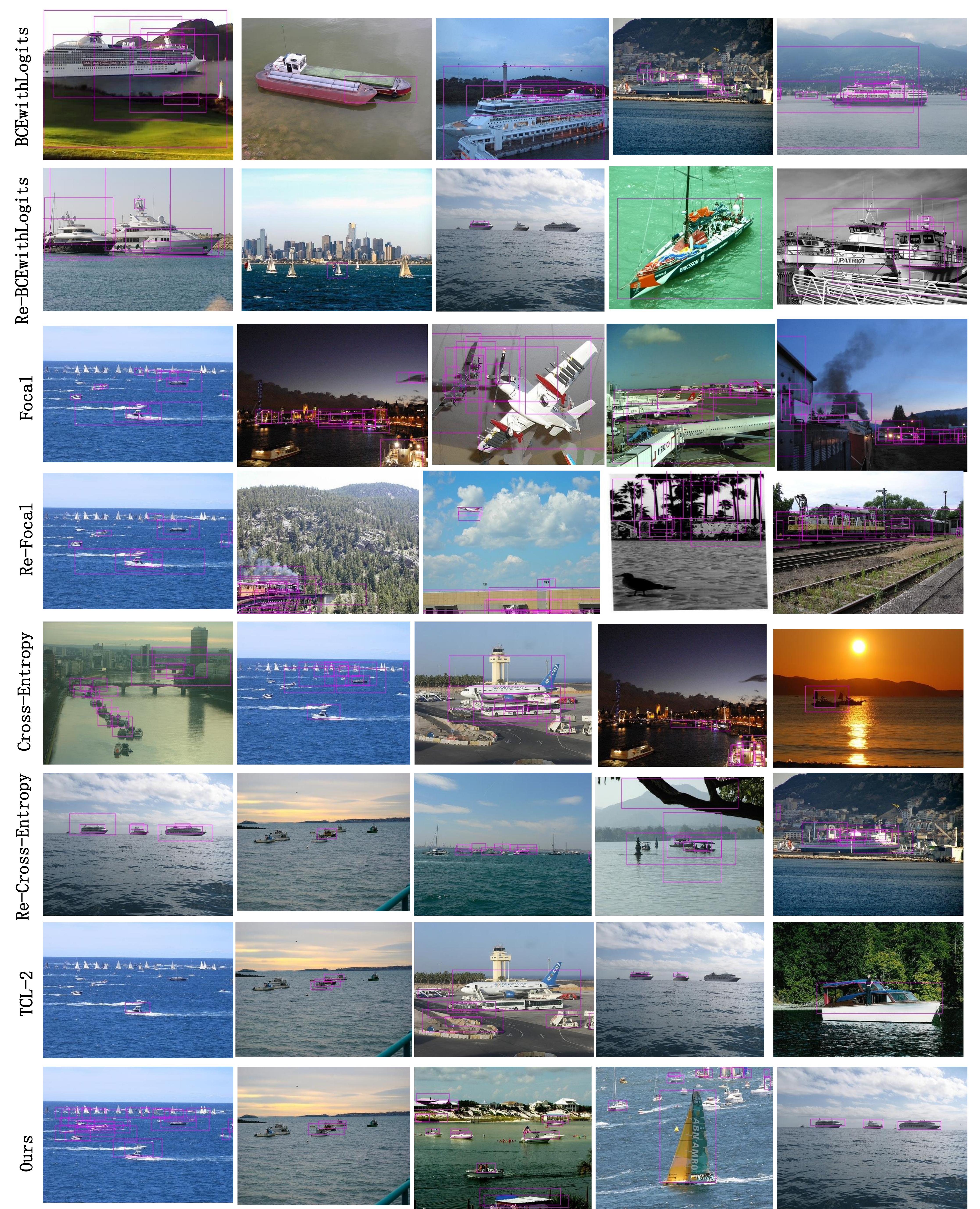}
  \caption{Qualitative 2-shot boat detection results on our test set for novel set1. We visualize the bounding boxes of all methods.}
  \label{show_boat}
\end{figure*}
\begin{figure*}
  \centering
  \includegraphics[width=1.0\columnwidth]{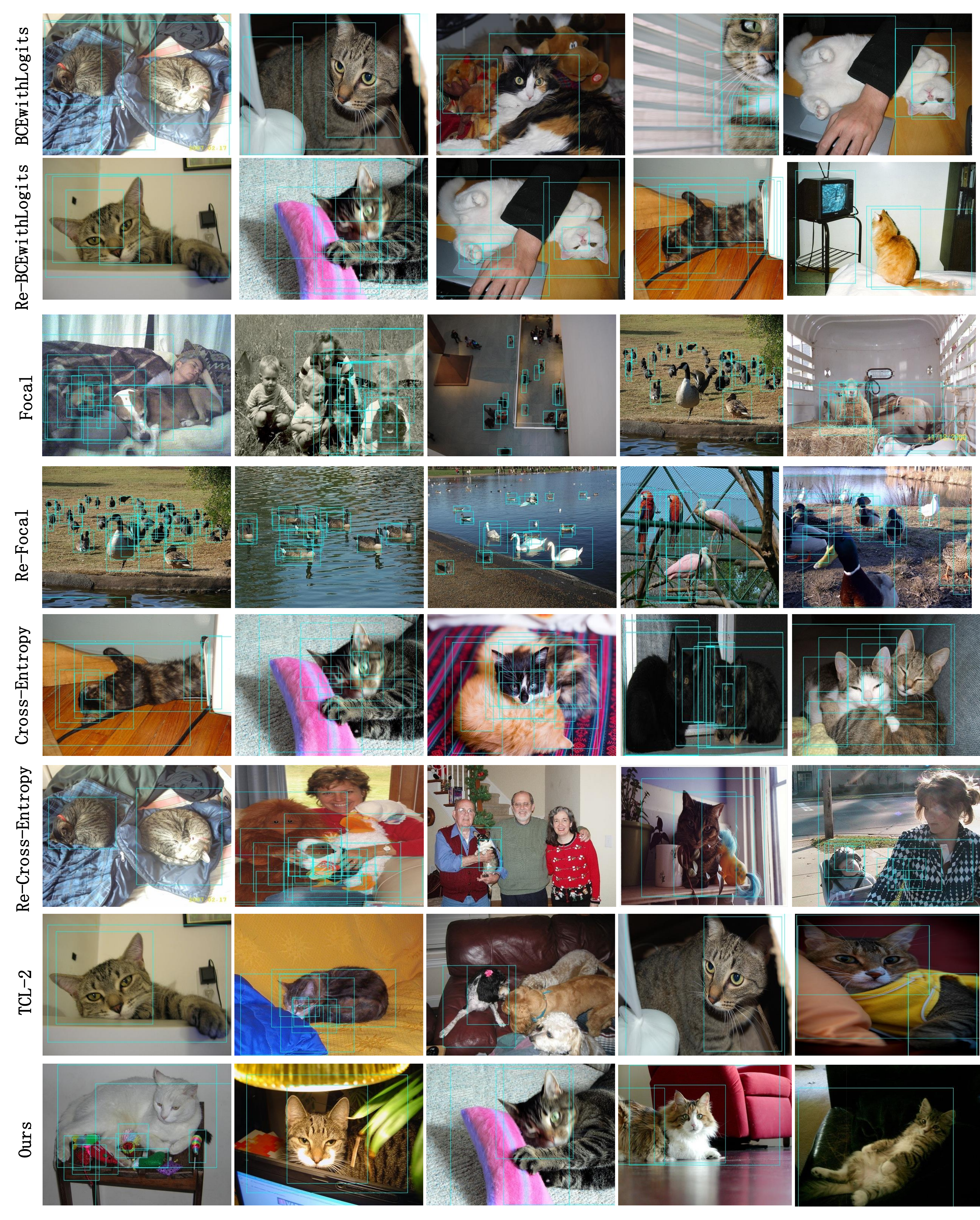}
  \caption{Qualitative 2-shot cat detection results on our test set for novel set1. We visualize the bounding boxes of all methods.}
  \label{show_cat}
\end{figure*}
\begin{figure*}
  \centering
  \includegraphics[width=0.85\columnwidth]{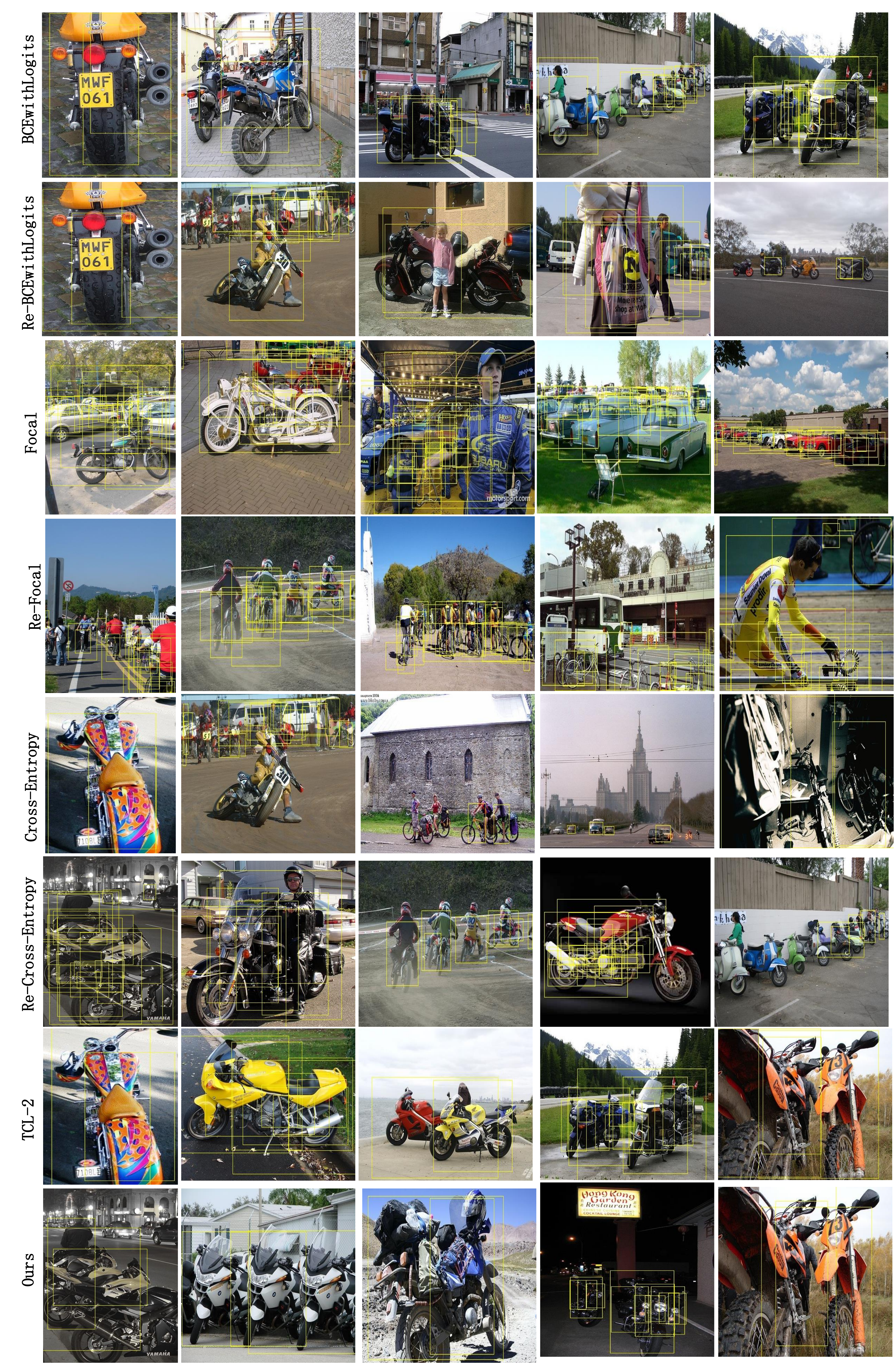}
  \caption{Qualitative 2-shot mbike detection results on our test set for novel set1. We visualize the bounding boxes of all methods.}
  \label{show_mbike}
\end{figure*}
\begin{figure*}
  \centering
  \includegraphics[width=0.85\columnwidth]{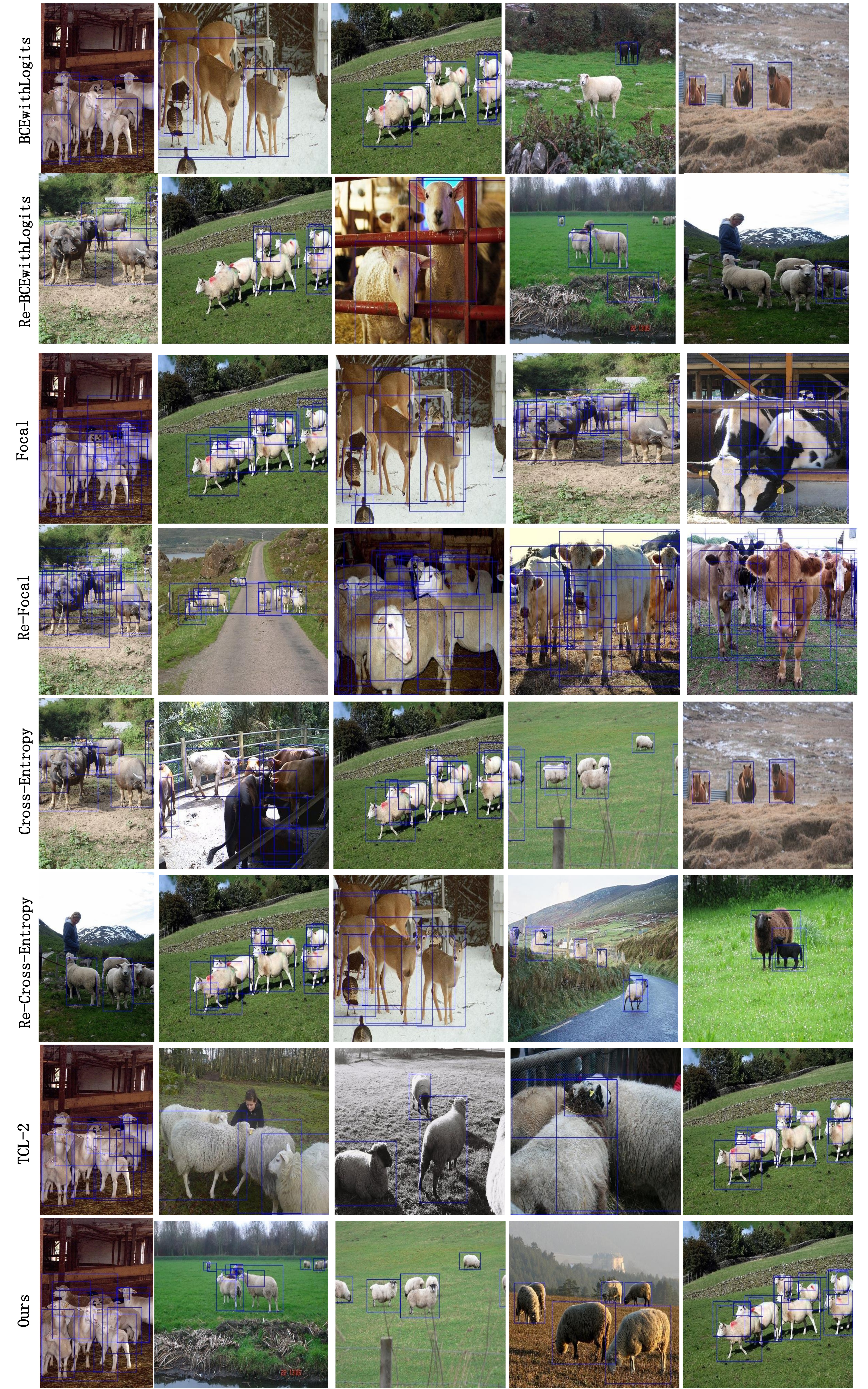}
  \caption{Qualitative 2-shot sheep detection results on our test set for novel set1. We visualize the bounding boxes of all methods.}
  \label{show_sheep}
\end{figure*}
\begin{figure*}
  \centering
  \includegraphics[width=1.0\columnwidth]{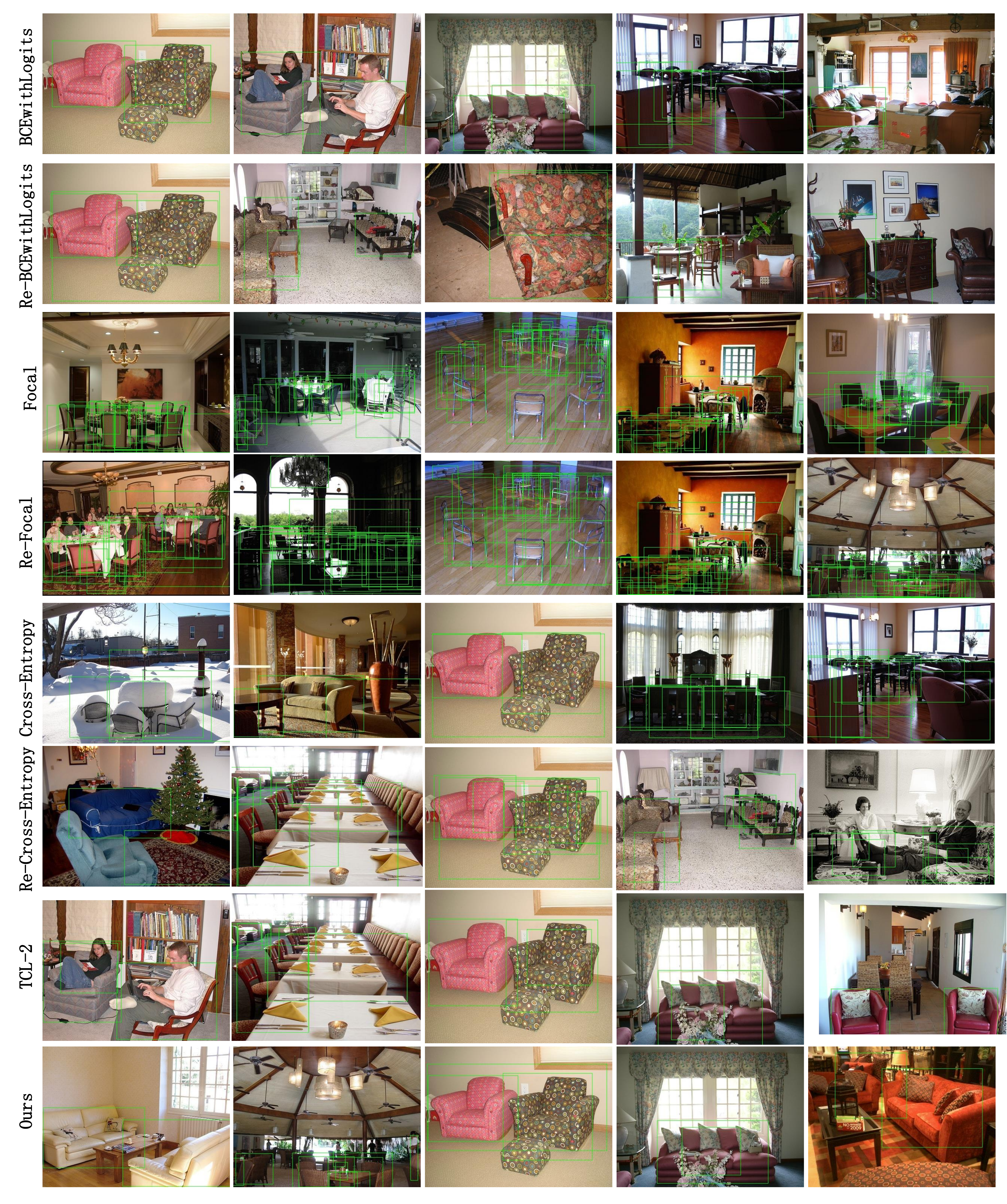}
  \caption{Qualitative 2-shot sofa detection results on our test set for novel set1. We visualize the bounding boxes of all methods.}
  \label{show_sofa}
\end{figure*}
\end{document}